\documentclass[runningheads]{llncs}
\usepackage{graphicx}
\usepackage{subcaption}
\usepackage{amsmath,amssymb} %
\usepackage{color}
\usepackage[width=122mm,left=12mm,paperwidth=146mm,height=193mm,top=12mm,paperheight=217mm]{geometry}
\usepackage{epsfig}
\usepackage{epstopdf}
\usepackage{multirow}
\usepackage{algorithm}
\usepackage{algorithmic}
\usepackage{cite}
\usepackage{tabularx, booktabs}
\usepackage{arydshln}

\newcolumntype{C}[1]{>{\centering\let\newline\\\arraybackslash\hspace{0pt}}m{#1}}
\newcolumntype{L}[1]{>{\raggedright\let\newline\\\arraybackslash\hspace{0pt}}m{#1}}

\makeatother

\usepackage{fancyheadings}

\usepackage{array,url,kantlipsum}
\newcommand{\Mark}[1]{\textsuperscript{#1}}

\usepackage[hang,flushmargin]{footmisc}

\newcolumntype{Y}{>{\centering\arraybackslash}X}

\newcommand\norm[1]{\left\lVert#1\right\rVert}
\begin{document}
\pagestyle{headings}
\mainmatter
\title{Self-Calibration Supported Robust Projective Structure-from-Motion} %

\title{Self-Calibration Supported Robust Projective Structure-from-Motion}
\titlerunning{Self-Calibration Supported Robust Projective SfM}
\authorrunning{Rui Gong, Danda Pani Paudel, Ajad Chhatkuli, and Luc Van Gool}
\author{Rui Gong\Mark{1}, Danda Pani Paudel\Mark{1}, Ajad Chhatkuli\Mark{1}, and Luc Van Gool\Mark{1,2}}
\institute{	\Mark{1}\,Computer Vision Lab, ETH Z\"urich, Switzerland \\
		\Mark{2}\,VISICS, ESAT/PSI, KU Leuven, Belgium \\
	\email{\{gongr,paudel,ajad.chhatkuli,vangool\}@vision.ee.ethz.ch}}
\maketitle

\begin{abstract}
Typical Structure-from-Motion (SfM) pipelines rely on finding correspondences
across images, recovering the projective structure of the observed
scene and upgrading it to a metric frame using camera self-calibration
constraints. Solving each problem is mainly carried out independently
from the others. For instance, camera self-calibration generally assumes
correct matches and a good projective reconstruction have been obtained.
In this paper, we propose a unified SfM method,
in which the matching process is supported by self-calibration constraints.
We use the idea that good matches should yield a valid calibration. In this process, we make use of the Dual Image of Absolute Quadric projection equations within a multiview correspondence framework, in order to obtain robust matching from a set of putative correspondences. The matching process classifies points as inliers or outliers, which is learned in an unsupervised manner using a deep neural network. Together with theoretical reasoning why the self-calibration constraints are necessary, we show experimental results demonstrating robust multiview matching and accurate camera calibration by exploiting these constraints. 
\keywords{Structure-from-Motion (SfM), Self-Calibration,  Dual Image of Absolute Quadric (DAQ)}
\end{abstract}

\section{Introduction}
Scene structure and camera motion recovery from uncalibrated images
is a fundamental problem in Computer Vision and a major requirement
for numerous three-dimensional capture systems. It has been established
that, in the absence of any knowledge about the scene and camera,
such structure can only be obtained up to a projective ambiguity.
As such reconstruction suffers from a severe distortion, it is only
useful in some limited applications (such as novel view synthesis).
In practice, most applications require the recovered projective scene
structure to be upgraded to metric. This upgrade is, however, not
achievable without the calibration of the camera. Traditionally, camera
calibration relies on information obtained from a known calibration
object present in the scene. Other methods rely on available measurements
directly extracted from the scene. One can argue that relying on scene
information may not be reliable as the assumed constraints might not
even be present in many cases.

With the development of flexible camera
calibration techniques \cite{zang,Tsai1987b}, cameras with fixed
intrinsic parameters can be reliably and accurately calibrated once
and used so long as the parameters are kept unchanged. If the known
calibration of the camera remains unchanged during the acquisition,
3D reconstruction boils down to solving linear systems of equations.
However, these parameters may very well vary before or during the
capture of the entire image sequence. The change may not take place
in every image captured but may, nevertheless, occur under change
of focus or zoom. As re-calibrating the camera in this fashion is
not always possible, it is safe to assume - in most cases - the camera
to be uncalibrated at any instant. Means to calibrate it, other than
relying on a special pattern or scene knowledge, are hence necessary.
One way to do so is to resort to the more advanced and flexible approach
of camera self-calibration, i.e. the recovery of the camera's parameters
using solely point correspondences across images. Point correspondences
across images, allow to locate a virtual object, the so-called Absolute
Conic (AC), that is omnipresent in all scenes. The AC is a special
conic lying on the plane at infinity and whose projection onto images
is independent upon the rigid motion of the camera. In particular,
the AC carries the advantage of projecting onto an image conic (IAC)
whose location only depends upon the intrinsic parameters of the camera
under consideration. Camera constraints, such as partial knowledge
or full parameters constancy, are used to fix the AC and its supporting
plane and hence calibrate the camera. This task is generally cast
into the problem of recovering a single object, the so-called Dual
Absolute Quadric (DAQ), encoding information about both the IAC (hence the
 camera intrinsics) and the AC's supporting plane (i.e. the
plane at infinity).

The recovery of the DAQ is a challenging nonlinear problem in which
correct correspondences are assumed to be available. Using point feature
detectors \cite{sift:99,surf:2008}, it is possible to extract
a good number of reliable features in an image. However, finding good
matches between the features obtained from two images of the same
scene is not an easy task. The problem becomes even more difficult
when it comes to simultaneously matching points across multiple images.
Matching based on the epipolar constraints is a widely used technique
for two views: given a point in one image, its corresponding point
in the second image lies on a known line. This constraint is not sufficient
enough to reject all the outliers as there may exist an outlier on
the other image that still lies on that line. When multiple images
are matched by considering only one pair of images at a time, there
may be a significant number of outliers. These outliers can further
be rejected by enforcing 3 or N views constraints. In practice, the
constraints only up to 3-views are used. This is due to the expensive
computational cost for higher number of views.

In general, feature extraction and matching are tackled independently
from the self-calibration problem. However, it is unrealistic to address
the camera self-calibration problem with the assumption of the availability
of perfectly matched sets of pixels across images. One can only notice
that, in order to facilitate the correspondence process, camera self-calibration
techniques are often tested only on ordered image sequences when real images
are considered. In the presence of mismatches, camera self-calibration
techniques are doomed to failure. Note that when camera parameters
are known, they can be used to support the matching of features through
the inspection of the re-projection residual errors. However, no such
approach exists when the calibration is unknown. The basic idea on
which our work is based upon spells out as follows: if self-calibration
was a linear process, then one could use it to support the correspondence
process in a way similar to the role of the fundamental matrix, the
trifocal (or multi-focal) matching tensors or even re-projection:
valid correspondences must necessarily lead to valid fundamental matrices,
valid multi-focal tensors and a valid re-projection of the reconstructed
scene. However, self-calibration constraints being inherently nonlinear,
they - seemingly - are of no use to support the search for correspondences
across uncalibrated images. We argue here that a packaged solution
of 3D reconstruction form multiview will not be complete unless both
of these problems are solved together. i.e, a robust multiview matching
and a reliable camera self-calibration that exploit one another instead
of being solved independently. How useful would be point correspondences
that yield inaccurate, false or even impossible calibration? Not useful
at all! Self-calibrating a camera with every candidate set of matches
is unrealistic and computationally prohibitive. Also, self-calibration
being a non-linear problem, it may very well fail because of numerical
optimization considerations rather than false matches. It is thus
of the utmost importance to find a proper formulation to express
the likely existence of a valid calibration (given a point-set match)
rather then going all the way to recover the camera parameters. The
main goal of this paper is specifically to match multi-view correspondences with the support of such intractable self-calibration constraints. Up to our knowledge, this is the first work that
(i) performs deep projective structure-from motion, and (ii) exploits the self-calibration constraints for the task of multi-view matching.

In this work, we design a deep unified framework for projective structure-from-motion and camera self-calibration, to support the multi-view matching process. Using a set of putative correspondences across multiple views, the proposed framework predicts inlier/outlier scores of the correspondences together with camera intrinsics and the plane-at-infinity. Notably, our deep network is trained end-to-end in an unsupervised manner. The unsupervised training for intrinsics and plane-at-infinity is possible, thanks to the self-calibration constraints expressed in the form of DAQ projection equations. In fact, when it is proven that our model is more robust and further improved by adding the self-calibration constraint when facing the difficult setting such as few points, few views and high outlier rate. We show the practicality of our methods, in terms of both robustness and accuracy, via real and the extreme cases of synthetic data.

\section{Related Works}

Camera self-calibration is widely known to be a difficult problem.
This is mainly due to two reasons: the nonlinear nature of the underlying
equations and the numerous critical motion sequences~\cite{CMS}.
Critical motions cause various levels of reconstruction ambiguities
and lead to the failure of camera calibration. The preliminary work
based on Kruppa's equation proposed in~\cite{faugeras92} is historically
seen as the first self-calibration method. However, its application
for three or more views provide weaker constraints than those obtained
through subsequent methods such as the one based on the modulus constraints~\cite{Pollefeys:96}
and the one relying on the Dual Absolute Quadric~\cite{triggs97}.
This is because Kruppa's constraints rely only on the dual images
of the AC and do not enforce that those images correspond to a unique
conic (the AC) lying on the plane at infinity~\cite{multiviewGeometry:2003}.
The plane at infinity and the AC are estimated in either of two ways;
one after another (stratified) or simultaneously (direct). The stratified
method given in \cite{KoenderinkandvanDoorn} for affine cameras was
extended to perspective in~\cite{faugeras95} with further developments
in~\cite{loung96, Pollefeys:96, adlakha2019quarch}.
Scene constraints combined with camera constraints
over multiple views is described in \cite{Liebowitz,pStrum99,Faugeras:95}.
The use of the modulus constraints to locate the plane at infinity
was introduced by Pollefeys in~\cite{Pollefeys:96}. 
The direct methods, which simultaneously estimate the plane at infinity and the
dual IAC, basically deal with DAQ~\cite{triggs97,habed2014efficient,rank3:2007}. The DAQ was
introduced for camera self-calibration by Triggs~\cite{triggs97}
who has proposed both quasi-linear and sequential programming methods
to locate it. Pollefeys et al.~\cite{Pollefeys-98} showed that the
DAQ computation could be used for metric reconstruction under general
motion even for varying focal length. In the case of a moving camera
with varying parameters, there exists no tight constraints on the
position of plane at infinity. Either chirality constraints~\cite{Nister:04}
or the finiteness constraint~\cite{Gherardi:2010} are used within
iterative search schemes. 
Stratified methods are sensitive to critical motions~\cite{multiviewGeometry:2003}. whereas, direct methods are and less problematic to critical sequences~\cite{pStrumFactorization:1996, gurdjos2009dual}, however are not flexible to be used in many cases.

All the above methods start with the common assumption on
the availability of perfect correspondences among all the images, which is not a practical. 
As per our knowledge, there is no
research work that simultaneously deals with multiview matching of
randomly captured images and self-calibration. 
One of the initial multiview matching work done for a set of unordered
real images is presented in~\cite{Schaffalitzky2002}. However, this
method does not find the correspondences among all the images. 
Recent methods for multi-view matching, although often in a different context, 
have also been developed~\cite{montserrat2019multi,schonberger2016pixelwise,serlin2019distributed}. On the other hand, almost all projective SfM  works~\cite{mahamud2001provably,hartley2003powerfactorization,oliensis2007iterative} are
primarily concerned for accuracy and optimality, without addressing the robustness, except two notable works that include~\cite{dai2010element, magerand2017practical}.
With the recent developments, learning-based methods for 3D reconstruction and/or self-calibration have also been developed~\cite{zhou2017unsupervised,godard2019digging,chen2019self, pedra2013camera, bogdan2018deepcalib, hold2018perceptual, gordon2019depth, zhuang2019degeneracy}. However, most of these works rely on the naive photo-metric error loss, if are unsupervised. This assumption immediately mandates ordered image sequence, or images captures under very similar conditions.
In regarding to learning-based matching for structure and/or motion, few notable works include~\cite{ranftl2018deep,probst2019unsupervised, brachmann2019neural,brachmann2017dsac}.

\section{Preliminaries}
The Fundamental matrix encapsulates all the necessary (projective) geometric relationships for the two-view imaging model. However, when more than two views are involved, more sophisticated relationships (analogous to the Fundamental matrix), involving measurements from all the views, are required. These relationships are known as $N$-view multilinear tensors such as the trifocal tensor for three views and the quadrifocal tensor for four. Although $N$-view tensors successfully encapsulate the geometric relationships upto 4 views, their usage is limited due to their computational complexities. Therefore, a common practice of incorporating measurements from multiple views involves the projective factorization method.  The process of projective factorization takes 2D point measurements from multiple views  and decomposes it into a scene structure and camera matrices that are consistent with this structure.

\subsection{Projective Factorization}\label{ssec:projFact}
Consider 3D points $\{X_{j}\}_{j=1}^m$ observed by cameras $\{P^{i}\}_{i=1}^n$. The observed image points are given by $\{{x}_{j}^{i}\}$. For given point correspondences $\{x_{j}^1 \leftrightarrow  x_{j}^i\}_{i=1}^n$ across images $\mathcal{I}^1,\mathcal{I}^2,\ldots,\mathcal{I}^n$, the reconstruction task is to find 3D point coordinates $\mathsf{X}_{j}$ and camera matrices $\mathsf{P}^{i}$ such that,
\begin{equation}
\mathsf{x}_{j}^{i}\sim\mathsf{P}^{i}\mathsf{X}_{j}, \,\,\,\,\,\mbox{for all } i \mbox{ and } j.\label{eq:pointProjection}
\end{equation}
If we write this equation explicitly by introducing scale variables (or Projective depth), we have,  $\lambda_j^i\mathsf{x}_{j}^{i}=\mathsf{P}^{i}\mathsf{X}_{j}$. Provided that the points are visible in all views (i.e.  $\mathsf{x}_{j}^{i}$ is known for all $i$ and $j$), the complete set of equations may be written by stacking the vectors and matrices in the following form,
  \begin{equation} \label{eq:projectiveFact1}
  {\arraycolsep=1.4pt\def\arraystretch{1.5}
\begin{bmatrix}
    \lambda_1^1\mathsf{x}_{1}^{1}	 & \lambda_2^1\mathsf{x}_{2}^{1}  & \dots  & \lambda_m^1\mathsf{x}_{m}^{1} \\
    \lambda_1^2\mathsf{x}_{1}^{2}    & \lambda_2^2\mathsf{x}_{2}^{2}  & \dots  & \lambda_m^2\mathsf{x}_{m}^{2} \\
    \vdots         					 & \vdots 				          & \ddots & \vdots \\
    \lambda_1^n\mathsf{x}_{1}^{n}    & \lambda_j^i\mathsf{x}_{2}^{n}  & \dots  & \lambda_m^n\mathsf{x}_{m}^{n}
\end{bmatrix} =
\begin{bmatrix}
    \mathsf{P}^{1} \\
    \mathsf{P}^{2} \\
    \vdots         \\
    \mathsf{P}^{n}
\end{bmatrix}
\begin{bmatrix}
    \mathsf{X}_{1} & \mathsf{X}_{2}   & \dots 	 & \mathsf{X}_{m} 
\end{bmatrix}.
}
\end{equation}
The matrix on the left-hand side is known as the measurement matrix, say $\mathsf{M}$. By construction, the matrix $\mathsf{M}$ is of rank 4.  This equation involves the scale variables $\lambda_j^i$, which are not part of the measurement, for each measured point $\mathsf{x}_j^i$. Furthermore, note that the decomposition on the right-hand side of the above equality is not unique. To see this, observe that with any non-singular $4\times 4$ matrix $\mathsf{H}$, we have $\mathsf{x}_{j}^{i}\sim\mathsf{P}^{i}\mathsf{H}^{-1}\mathsf{H}\mathsf{X}_{j}$ which is also satisfied. Such reconstruction $\{P^{i},X_{j}\}$ is a projective reconstruction and the matrix $\mathsf{H}$ is called a projective homography matrix. There are several approaches that allow decomposing the measurement matrix $\mathsf{M}$ in the form of Equation~(\ref{eq:projectiveFact1}).

{\bf Sturm/Triggs Factorization:} The first solution to decomposed $\mathsf{M}$  as ~(\ref{eq:projectiveFact1}) was proposed by {\it Sturm and Triggs}~\cite{sturm1996factorization}, where the initial estimate of projective depths $\lambda_j^i$ is assumed to be known. This may be obtained either from initial projective reconstruction (for example, using fundamental matrix) or simply setting all $\lambda_j^i = 1$. Once the projective depths are known, the measurement matrix  $\mathsf{M}$ is complete. In case of noisy measurements, the $\mathsf{M}$ can be enforced to have rank 4 using Singular Value Decomposition.  Thus, if $\mathsf{M} = \mathsf{UD}\mathsf{V}^T$, all except the largest four diagonal entries of $\mathsf{D}$ are forced to zero resulting in $\hat{\mathsf{D}}$. Then, the rank constrained measurement matrix is $\mathsf{M} = \mathsf{U}\hat{\mathsf{D}}\mathsf{V}^T$. Using such decomposition, the camera matrices and the scene points are retrieved as,
\begin{align}
\begin{bmatrix}
    \mathsf{P}^{1T} & \mathsf{P}^{2T} &  \ldots  & \mathsf{P}^{nT}
\end{bmatrix}^T = \mathsf{U}\hat{\mathsf{D}}^{\frac{1}{2}}, \nonumber
\\
\begin{bmatrix}
    \mathsf{X}_{1} & \mathsf{X}_{2}   & \dots 	 & \mathsf{X}_{m} \\
\end{bmatrix} =  \hat{\mathsf{D}}^{\frac{1}{2}}\mathsf{V}^T.
\end{align}

\subsection{Projective-to-Metric Upgrade}
For simplicity and without loss of generality, we assume that the coordinate frame of the first camera in both projective and metric space coincide with the world frame such that the first cameras are respectively given by, $\mathsf{P}^{1}=\mathsf{[I\,|\,0]}$ and $\mathsf{P}^{1}_{\mathsf{M}}=\mathsf{K}^1\mathsf{[I\,|\,0]}$. The projective structure and motion can then be upgraded using,
\begin{align}
\mathsf{P}^{i}_{\mathsf{M}} \sim \mathsf{P}^{i}\mathsf{H}_{\mathsf{M}}^{-1} \mbox{ and }\, \mathsf{X}_{j}^{\mathsf{M}} \sim \mathsf{H}_{\mathsf{M}}\mathsf{X}_{j}  \,\,\,\,\,\mbox{for all } i \mbox{ and } j, \text{ with,} \nonumber
\\
\mathsf{H}_\mathsf{M} \sim \left[\begin{array}{cc}                     {\mathsf{K}^1}^{-1} & {0}\\ 
   n_\infty^{T} & 1
   \end{array}\right]  \mbox{ and }\,
\mathsf{H}_\mathsf{M}^{-1} \sim \left[\begin{array}{cc}
   \mathsf{K}^1 & 0\\
   -n_\infty^{T}\mathsf{K}^1 & 1
   \end{array}\right], 
\label{eq:projMatProjeAndEuclid}
\end{align}
where the $\mathsf{\Pi}_\infty = (n_\infty^\intercal \,1)^\intercal$ are the coordinates of the so-called plane at infinity, say $\Pi_\infty$, in the projective space whose frame coincides to the first camera.

\subsection{DAQ for Camera Self-Calibration} \label{subsec:DAQ_selfcalib}
The Dual Absolute Quadric (DAQ), $\mathsf{Q_{M}}$, is a special
degenerate quadric of planes in the dual 3D-space \cite{multiviewGeometry:2003}.
The canonical form of DAQ in metric space is given
by, ${\mathsf{Q_M} = \bigl(\begin{smallmatrix}\mathsf{I}&0 \\ 0&0
\end{smallmatrix} \bigr)}$, which is fixed under metric transformations and takes the form 
$\mathsf{Q} = \mathsf{H}_\mathsf{M}^{-1} \bigl(\begin{smallmatrix}\mathsf{I}&0
 \\ 0&0 \end{smallmatrix} \bigr)\mathsf{H}_\mathsf{M}^{-\intercal}$ 
in the projective space. Using the form of $\mathsf{H}_\mathsf{M}^{-1}$ in~\eqref{eq:projMatProjeAndEuclid}, one can express DAQ in projective space with respect to the first camera frame as,
\begin{equation}
    \mathsf{Q= 
\left[\begin{array}{cc}
\omega^{1} & -\omega^{1}n_{\infty}^{T}\\
-n_{\infty}\omega^{1} & \,\, n_{\infty}\omega^{1}n_{\infty}^{T}
\end{array}\right]} \text{ with } \mathsf{\omega^1={K}^1{(K^1)^\intercal}}, 
\label{eq:DAQform}
\end{equation}
where, $\mathsf{\omega^1}$ is also known as the Dual Image of Absolute Conic (DIAC) in the first image. Direct self-calibration methods rely on the existence of DAQ of the form~\eqref{eq:DAQform}. More specifically, one can establish the relationships between DAQ and DIAC in each view using the projective projection matrices as follows, 
\begin{equation}
\mathsf{\omega}^i\sim \mathsf{P}^i\mathsf{Q}(\mathsf{P}^{i})^\intercal.
\label{eq:DAQ_projection}
\end{equation}
In this regard, the task of self-calibration is finding ${\mathsf{Q}\in \mathbb
{R}^{4 \times 4}}$ that has structure of~\eqref{eq:DAQform} and satisfies~\eqref{eq:DAQ_projection}, using the given projective projection matrices.

\section{Mulitview Matching}
The process of multiview matching assumes that the putative correspondences may get contaminated by potentially overwhelmingly many outlying matches. The filtering of these outliers is carried out while maximizing the consensus set of the correspondences that respect the factorization process of~\eqref{eq:projectiveFact1}, while respecting the DAQ projection of~\eqref{eq:DAQ_projection}. In this process, we are interested on classifying correspondence with the help of $\mathsf{w}=\{0,1\}^m$,  for given noise free outlier contaminated measurement matrix $\mathsf{M}$, using the following optimization problem, 
\begin{equation}
{\begin{array}{ll}
\displaystyle \max_{\substack{\mathsf{w,X,P, Q}}}	&	\sum_j{w_j}\\
\mbox{subject to}& \text{diag}(\mathsf{w})\odot\mathsf{(M-PX)} = 0,\\
& \mathsf{P}^i\mathsf{Q}(\mathsf{P}^{i})^\intercal \sim \mathsf{\omega}^i,\,\,\,\,\, \text{for}\,\,\, i=1,\ldots,n,\end{array}}
\label{eq:optimizationProblem1}
\end{equation}
where, $\odot$ denotes the element-wise matrix multiplication, $\mathsf{X,P,Q}$ are  projective structure, motion, and DAQ, respectively. Note that the assignment variable $w_i=1$ implies that the measurement corresponding to the point $X_i$ is an inliers, otherwise it is an outlier. The optimization problem of~\eqref{eq:optimizationProblem1}, however, has two major issues that need to be addressed prior to be used in practice. One concerns about noise and the other about an efficient usage of DAQ projection equation. 

\subsection{In the Presence of Noise} \label{sec:rank_constraint}
When the measurement matrix $\mathsf{M}$ is also contaminated by noise, the constraint $\text{diag}(\mathsf{w})\odot\mathsf{(M-PX)} = 0$ of~\eqref{eq:optimizationProblem1} is not often satisfied for the desired solution.  Therefore, we instead seek for a matrix ${\mathsf{S}\in\mathbb{R}^{3n\times m}}$, which is closest, in Frobenius norm, to  the outlier filtered measurement matrix by ensuring the following,
\begin{equation}\label{eq:noiseReplacement}
{\begin{array}{ll}
\displaystyle \min_{\substack{\mathsf{S}}}	&	\norm{\mathsf{S-\text{diag}(\mathsf{w})\odot\mathsf{M}}},\\
\mbox{subject to}& \mathsf{S}-\text{diag}(\mathsf{w})\odot\mathsf{PX}=0.\end{array}}
\end{equation}
In fact, any $\mathsf{S}\in\mathbb{R}^{3n\times m}$ with $\text{rank}(\mathsf{S})=4$ satisfies the constraint of~\eqref{eq:noiseReplacement}. On the other hand, the rank-4 matrix that minimizes the objective of~\eqref{eq:noiseReplacement} can be obtained by using the singular value decomposition of $\text{diag}(\mathsf{w})\odot\mathsf{M}$, whenever the assignment variable $\mathsf{w}$ is known. Note that any matrix of higher rank can be projected on the rank-4 manifold by setting all except largest four singular values to zero, similar to the Sturm/Triggs Factorization  discussed in Section~\ref{ssec:projFact}. 
 
\subsection{DAQ Projection for Constant Intrinsics}  \label{sec:DAQ_assump}
The DAQ projection constraint of~\eqref{eq:DAQ_projection} may turn out to be weak, if we assume that each camera can have different intrinsics. This however, is not a problem in itself. One can still make use of the DAQ projection constraints under the known prior in intrinsics. The known prior may include zero skew, unit aspect ratio, principal point close to image center, or only change in focal lengths.
In this work, we assume that all the cameras have constant intrinsics. Furthermore, we also need to consider that projection equation will not be satisfied exactly in the presence of noise. Therefore, we minimize the following objective function,
\begin{equation} \label{eq:DAQ_constraint}
    \mathcal{\eta}(\mathsf{Q}) = \sum_i{ \norm{\frac{\mathsf{P}^i\mathsf{Q}(\mathsf{P}^{i})^\intercal}{\norm{\mathsf{P}^i\mathsf{Q}(\mathsf{P}^{i})^\intercal}}-\frac{\mathsf{\omega}^1}{\norm{\mathsf{\omega}^1}}}}.
\end{equation}

\subsection{The Matching Objective}
The primarily goal of the multiview matching  is to compute inlier/outlier assignment that also satisfy the  DAQ projection conditions.  Therefore, we aim at simultaneously estimating $\mathsf{w}$ and $\mathsf{Q}$ by maximizing the surrogate objective, of~\eqref{eq:optimizationProblem1}, stated as follow,   
\begin{equation}
{
\mathcal{L}(\mathsf{w},\mathsf{Q}) = 	-\sum_j{w_j}+\alpha\sum_{k=5}^m{\sigma_k( \text{diag}(\mathsf{w})\odot\mathsf{M})}+\beta\mathcal{\eta}(\mathsf{Q}),
\label{eq:optimizationProblemFinalLoss}
}\end{equation}
where $\alpha>0, \beta>0$ are the weights that take care of the influence of noise and constant intrinsics factors, respectively.

\subsection{Self-Calibrating Projective SfM} \label{sec:SCSFM}
In order to solve the optimization problem of~(\ref{eq:optimizationProblemFinalLoss}), we present our deep self-calibrating projective SfM model (SCPSfM), which simultaneously performs the projective factorization and camera calibration. In this process, we exploit the advantage of a deep neural network on the high optimization accuracy and efficiency. Starting from noise and outliers contaminated measurement matrix $\mathsf{M}$, defined in Section~\ref{ssec:projFact}, we predict per correspondence weights $\mathcal{W}=\{w_{j}\in [0,1]\}$, defined in~\ref{eq:optimizationProblem1}), to detect the inliers ($w_{j}\rightarrow 1$) and outliers ($w_{j}\rightarrow 0$) correspondence. Additionally, we also accurately calibrates the camera intrinsics $\mathsf{K}$ defined in~(\ref{eq:projMatProjeAndEuclid}) and the coordinate of the plane at infinity $\mathsf{n}_\infty$ defined in~(\ref{eq:DAQform}). Let us denote our SCPSfM model as $S_{\theta}:\mathsf{M}\rightarrow \mathsf{w}\times (\mathsf{K}, \mathsf{n}_{\infty})$, which parameterized by $\theta$. Using the measurement matrix $\mathsf{M}$ as input, our SCPSfM model predicts inlier/outlier scores as well as self-calibrates the camera, without requiring any ground truth, whatsoever. SCPSfM relies on DAQ projection and projective factorization constraints presented in~(\ref{eq:noiseReplacement}) and (\ref{eq:DAQ_constraint}). Based on the objective function of~(\ref{eq:optimizationProblemFinalLoss}), we propose the total loss function $\mathcal{L}(\theta, \mathsf{M})$ which combines the projection assignment loss $\mathcal{L}_{proj}(\theta, \mathsf{M})$, and the DAQ loss $\mathcal{L}_{DAQ}(\theta, \mathsf{M})$, and the inlier loss $\mathcal{L}_{num}(\theta, \mathsf{M})$ resulting the following total loss:
\begin{equation} \label{eq:total_loss}
\begin{aligned}
    \mathcal{L}(\theta, \mathsf{M}) =& \mathcal{L}_{num}(\theta, \mathsf{M}) + \alpha \mathcal{L}_{proj}(\theta, \mathsf{M})+\beta \mathcal{L}_{DAQ}(\theta, \mathsf{M}), \,\,\,\,\,\text{with,}\\
    \mathcal{L}_{num}(\theta, \mathsf{M}) &= exp(t-\sum_{j=1}^{n}s(w_{j}(\theta, \mathsf{M})-0.5)), \\
    \mathcal{L}_{proj}(\theta, \mathsf{M}) &= \sigma_4( \text{diag}(\mathsf{w}(\theta, \mathsf{M}))\odot\mathsf{M}), \\
    \mathcal{L}_{DAQ}(\theta, \mathsf{M}) &= \sum_i{ \norm{\frac{\mathsf{P}^i\mathsf{Q}(\mathsf{P}^{i})^\intercal}{\norm{\mathsf{P}^i\mathsf{Q}(\mathsf{P}^{i})^\intercal}}-\frac{\mathsf{\omega}^1}{\norm{\mathsf{\omega}^1}}}}, \\
\end{aligned}
\end{equation}
where $\alpha$ and $\beta$ are hyper parameters which balance between different loss, $s(\cdot)$ represents the sigmoid function.  The hyper parameter $t$ represents threshold that guarantees the least number of inliers detected. The projection matrices $\mathsf{P}^i$ can be recovered from $\text{diag}(\mathsf{w}(\theta, \mathsf{M}))\odot\mathsf{M}$ according to~(\ref{eq:projectiveFact1}),  whereas the DAQ $\mathsf{Q}$ can be derived using $(\mathsf{n}_\infty(\theta, \mathsf{M}), K(\mathsf{M}))$ based on~(\ref{eq:DAQform}). 

Our input is the measurement matrix, where each correspondence can been seen as a point in $3n$-dimensional space. Our  output aims to assign a label of inlier or outlier to each of the correspondence. This problem can be seen as a one-class point segmentation problem. A typical networks for point cloud segmentation naturally meets our requirement.  Therefore, we use adopt PointNet~\cite{qi2017pointnet} as the basic building block of our SCPSfM model. 

\begin{figure*}
    \centering
    \includegraphics[width=\textwidth]{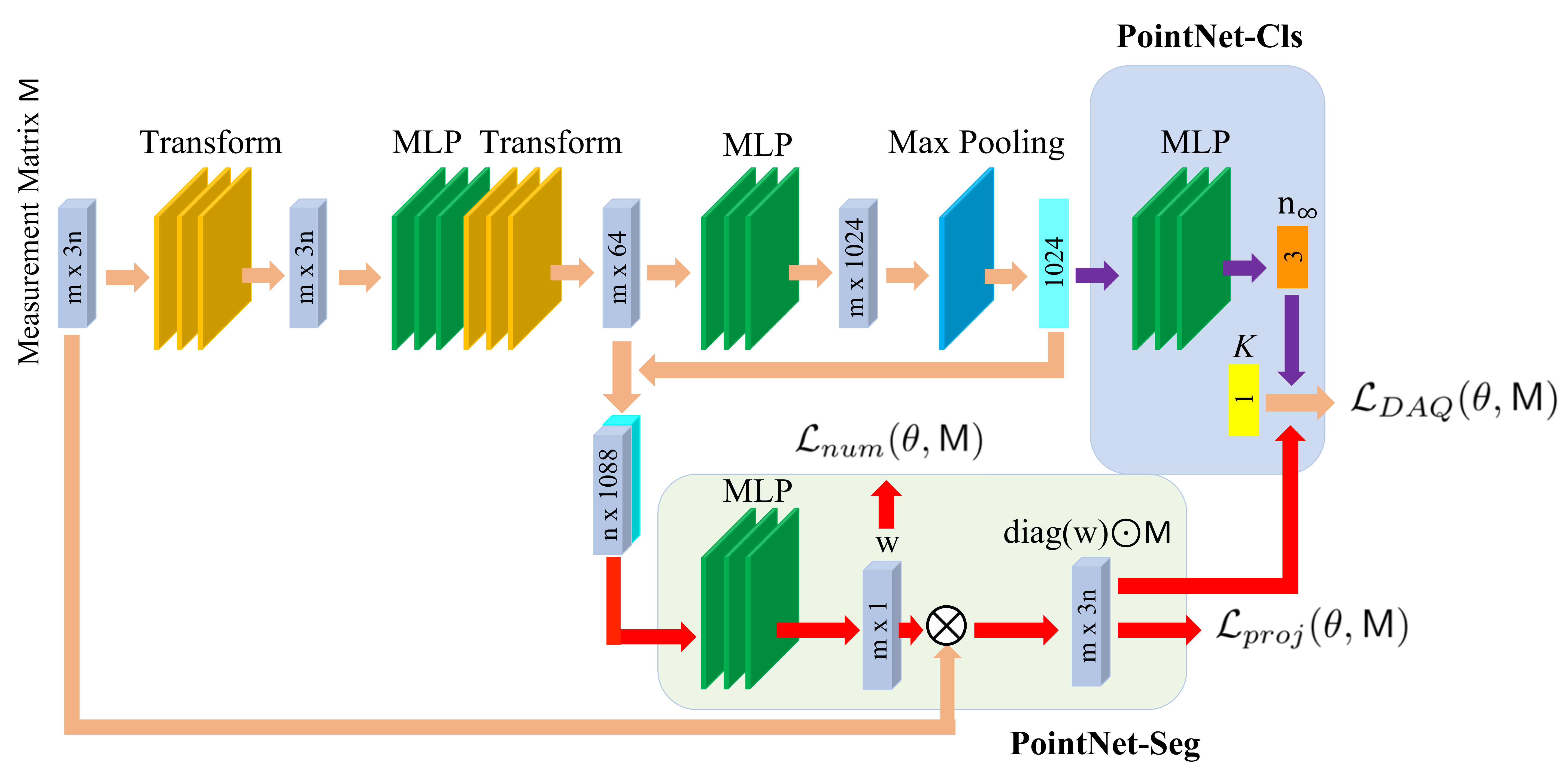}
    \caption{The overview of our SCPSfM model: the PointNet-Seg and PointNet-Cls share the same feature extraction part before the 1024-dim global feature extraction. The Point-Seg branch predicts the correspondence weight $w_j$ for the $j$-th correspondence to distinguish inlier/outlier. ,The Point-Cls branch regresses the plane at infinity $\mathsf{n}_{\infty}$. Besides, camera intrinsics $K$ are set as  network parameters, as a part of $\theta$. The  purple arrow indicates the unique flow of the PointNet-Cls branch. Similarly, the red arrow denotes the unique flow of the PointNet-Seg branch, while the orange one represents the common flow shared by both.}
    \label{fig:network_structure}
\end{figure*}

\textbf{Implementation.} We illustrate our network structure in Fig. \ref{fig:network_structure}. Based on the theory presented in Section \ref{sec:SCSFM}, we denote the SCPSfM model as the mapping $S_{\theta}:\mathsf{M}\rightarrow \mathsf{w}\times (K, \mathsf{n}_\infty)$. Then the network structure of our SCPSfM model can be divided into two branches, one assigns weight $w_{j}$ to each of the correspondence and the other regresses the  plane at infinity $\mathsf{n}_{\infty}$. In order to ensure constant intrinsics $K$, as stated in Section \ref{sec:DAQ_assump}, the camera intrinsics are set as network parameters, a part of $\theta$.  As shown in Fig. \ref{fig:network_structure}, SCPSfM model combines PointNet-Seg and PointNet-Cls structures to realize the two branches. The two branches share the common part for feature extraction. The PointNet-Seg branch outputs the $m$-dimensional vector $\mathsf{w} = [w_1, \dots, w_m]^\intercal$ for inlier/outlier scores of the correspondences. The PointNet-Cls branch regresses the 3-dimensional coordinate of the plane at infinity $\mathsf{n}_{\infty}$. We implement the SCPSfM model in Pytorch \cite{NEURIPS2019_9015} and use the ADAM \cite{kingma2014adam} optimizer to train the network.

\begin{figure*}
\centering
\begin{subfigure}[c]{0.48\textwidth}
  \centering
  \includegraphics[width=\linewidth]{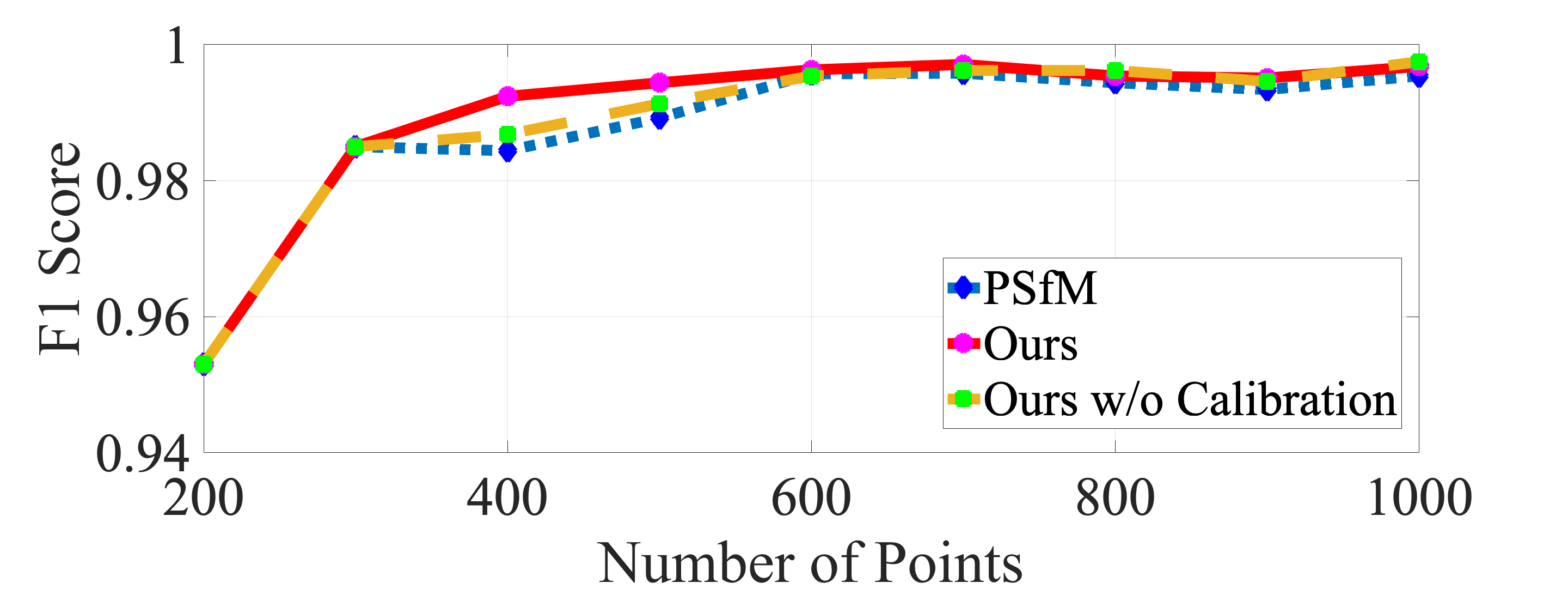}
  \subcaption{$n=20, \sigma = 0.6\%, \delta = 0.9$, $m$ varies}
  \label{syn_comp_F1:sfig1}
\end{subfigure}
\begin{subfigure}[c]{0.48\textwidth}
  \centering
  \includegraphics[width=\linewidth]{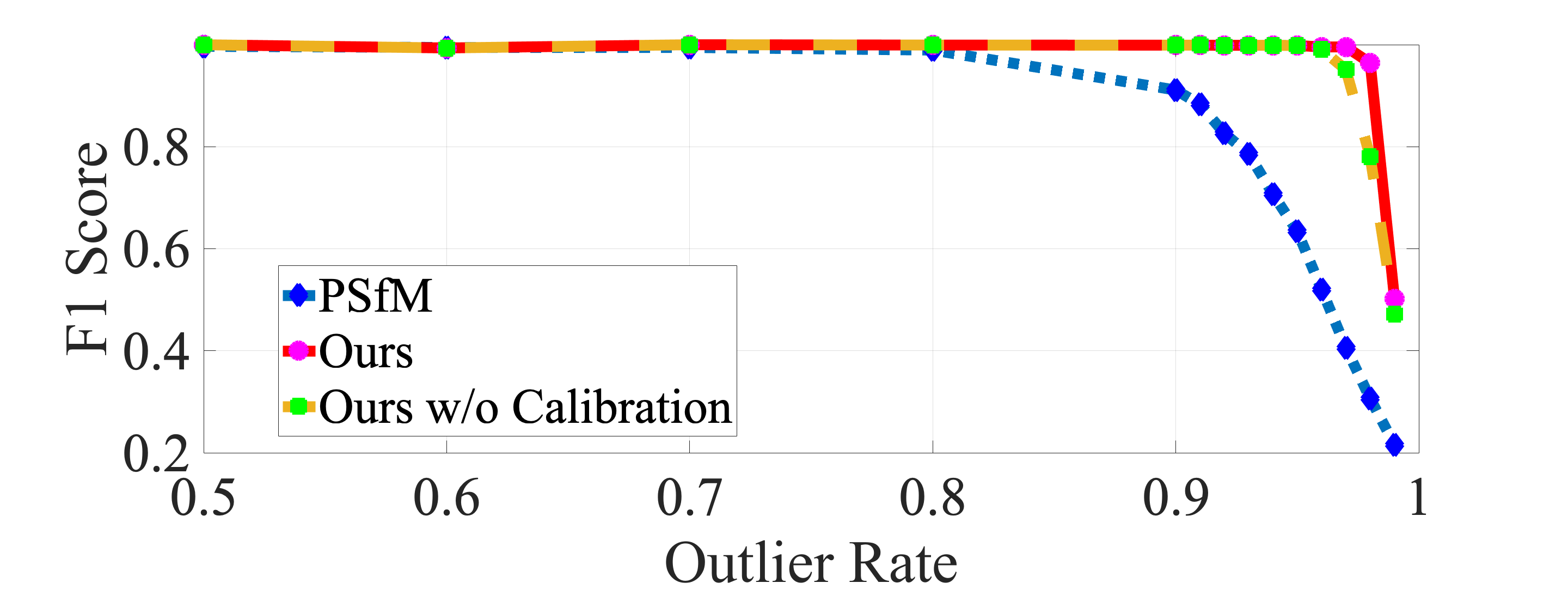}
  \subcaption{$n=10, m=200, \sigma=0.3\%$, $\delta$ varies}
  \label{syn_comp_F1:sfig2}
\end{subfigure}
\begin{subfigure}[c]{0.48\textwidth}
  \centering
  \includegraphics[width=\linewidth]{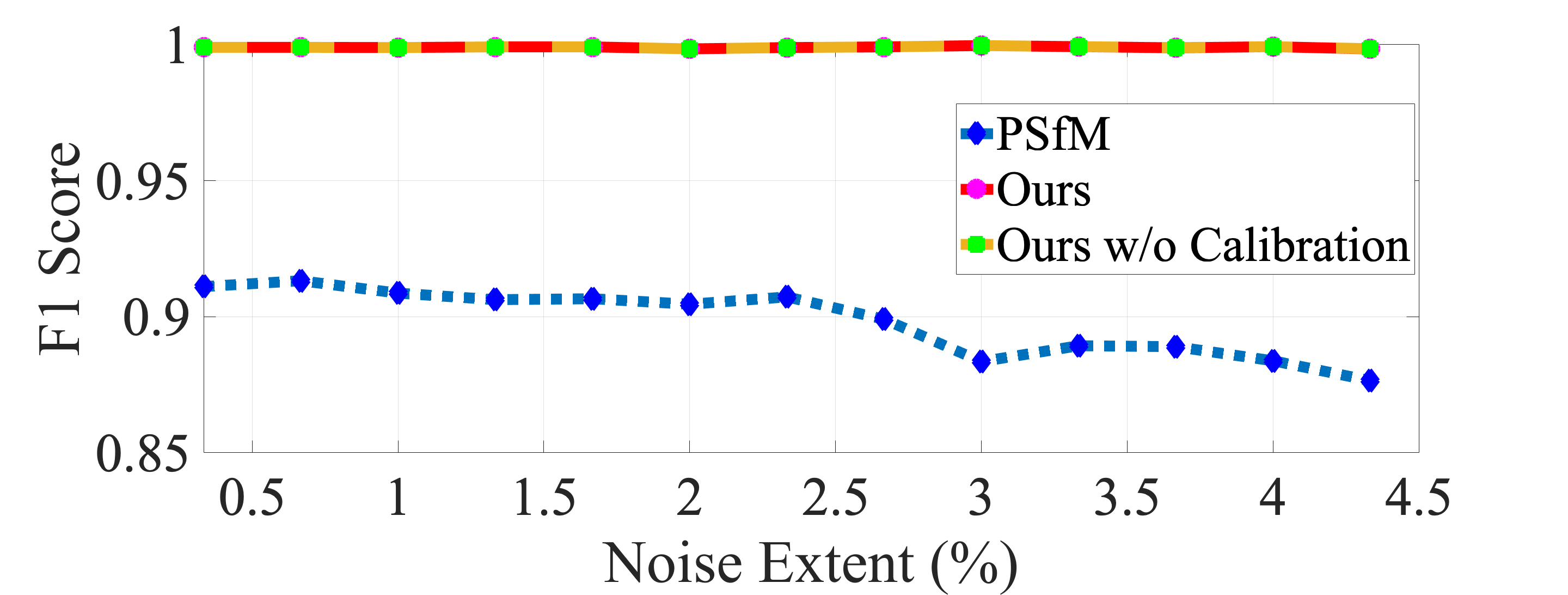}
  \subcaption{$n=10, m=200, \delta=0.9$, $\sigma$ varies}
  \label{syn_comp_F1:sfig3}
\end{subfigure}
\begin{subfigure}[c]{0.48\textwidth}
  \centering
  \includegraphics[width=\linewidth]{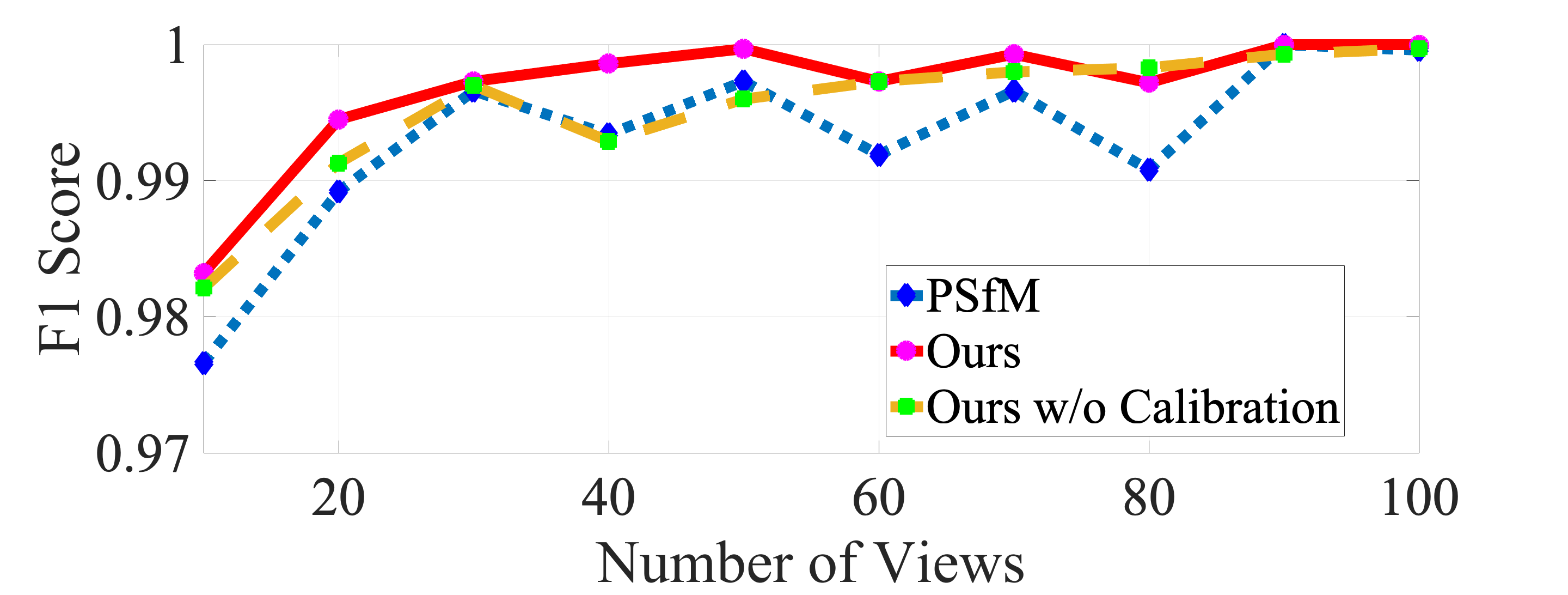}
  \subcaption{$m=500, \sigma = 0.6\%, \delta = 0.9$, $n$ varies}
  \label{syn_comp_F1:sfig4}
\end{subfigure}
\begin{subfigure}[c]{0.48\textwidth}
  \centering
  \includegraphics[width=\linewidth]{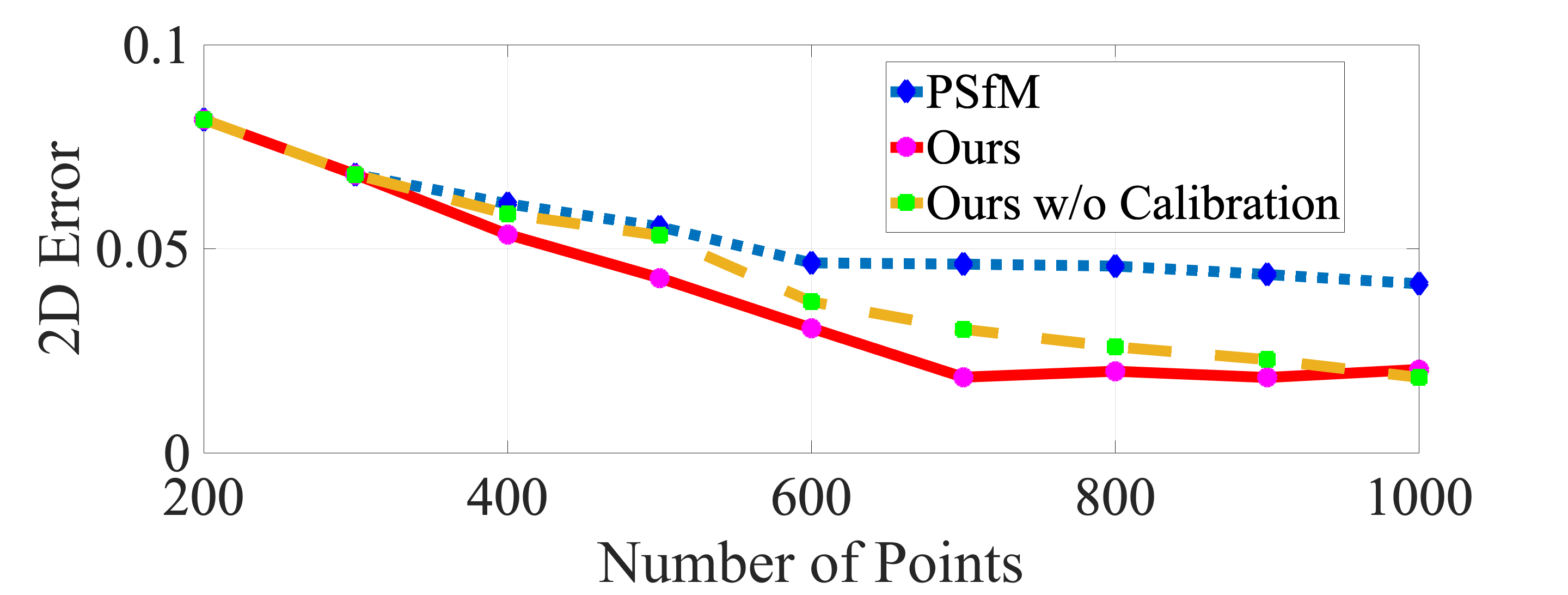}
  \caption{$n=20, \sigma = 0.6\%, \delta = 0.9$, $m$ varies}
  \label{syn_comp_2D:sfig1}
\end{subfigure}
\begin{subfigure}[c]{0.48\textwidth}
  \centering
  \includegraphics[width=\linewidth]{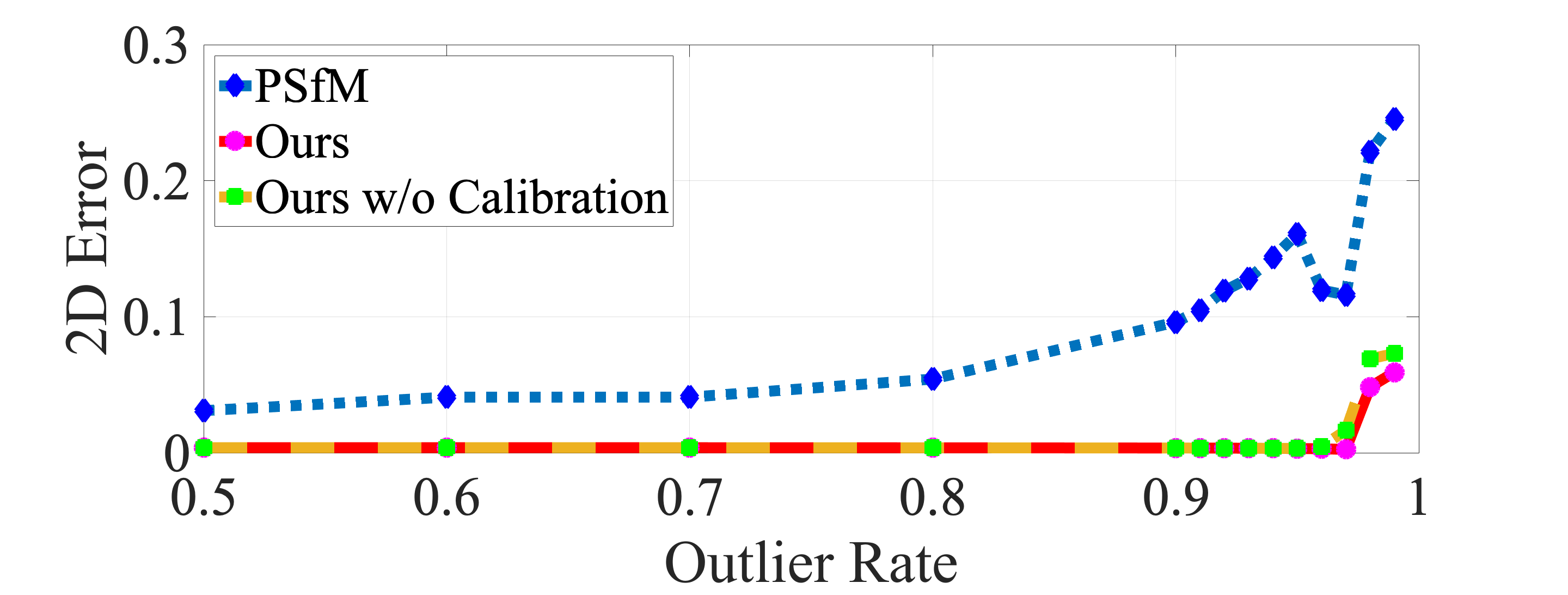}
  \caption{$n=10, m=200, \sigma=0.3\%$, $\delta$ varies}
  \label{syn_comp_2D:sfig2}
\end{subfigure}
\begin{subfigure}[c]{0.48\textwidth}
  \centering
  \includegraphics[width=\linewidth]{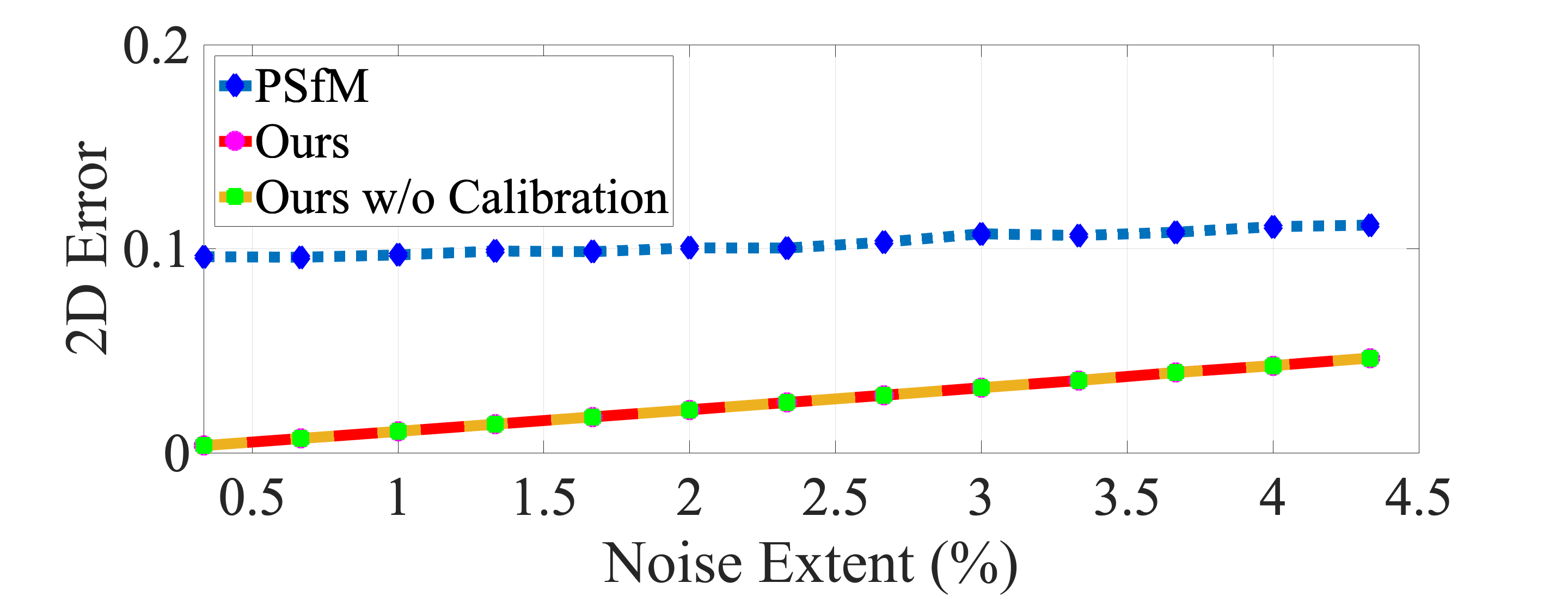}
  \caption{$n=10, m=200, \delta=0.9$, $\sigma$ varies}
  \label{syn_comp_2D:sfig3}
\end{subfigure}
\begin{subfigure}[c]{0.48\textwidth}
  \centering
  \includegraphics[width=\linewidth]{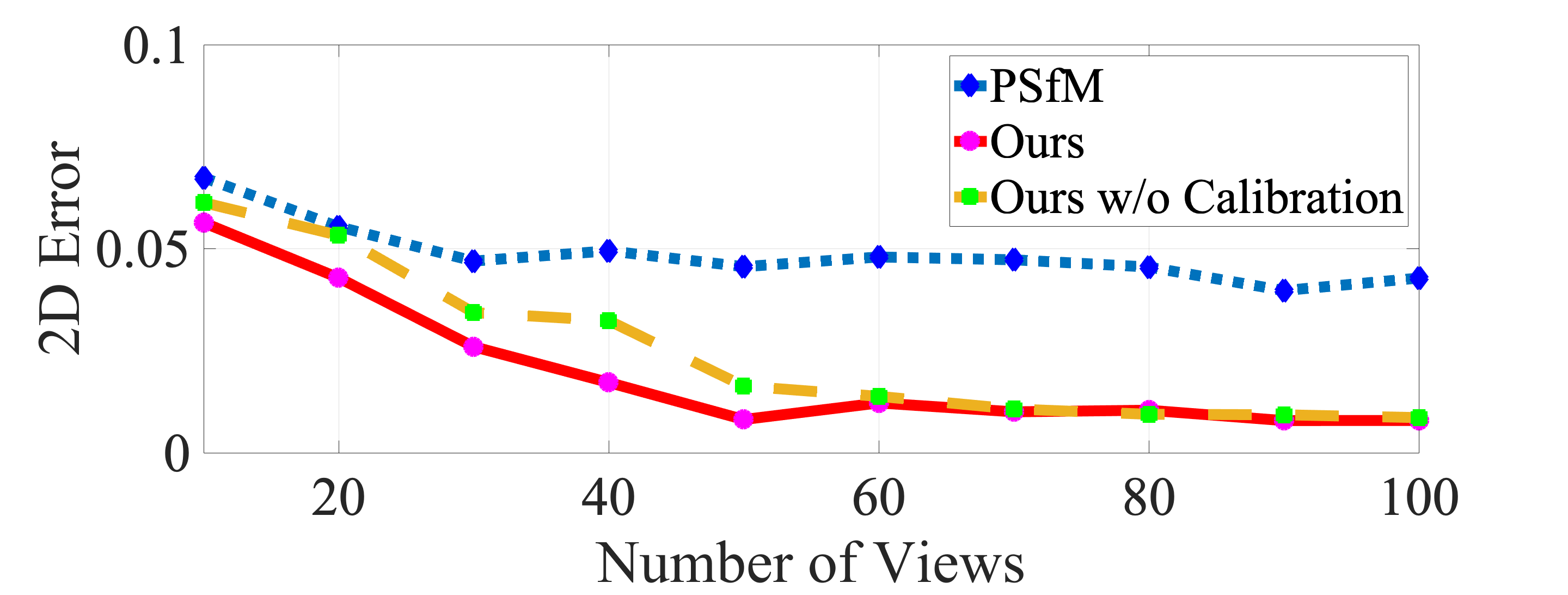}
  \caption{$m=500, \sigma = 0.6\%, \delta = 0.9$, $n$ varies}
  \label{syn_comp_2D:sfig4}
\end{subfigure}
\begin{subfigure}[c]{0.48\textwidth}
  \centering
  \includegraphics[width=\linewidth]{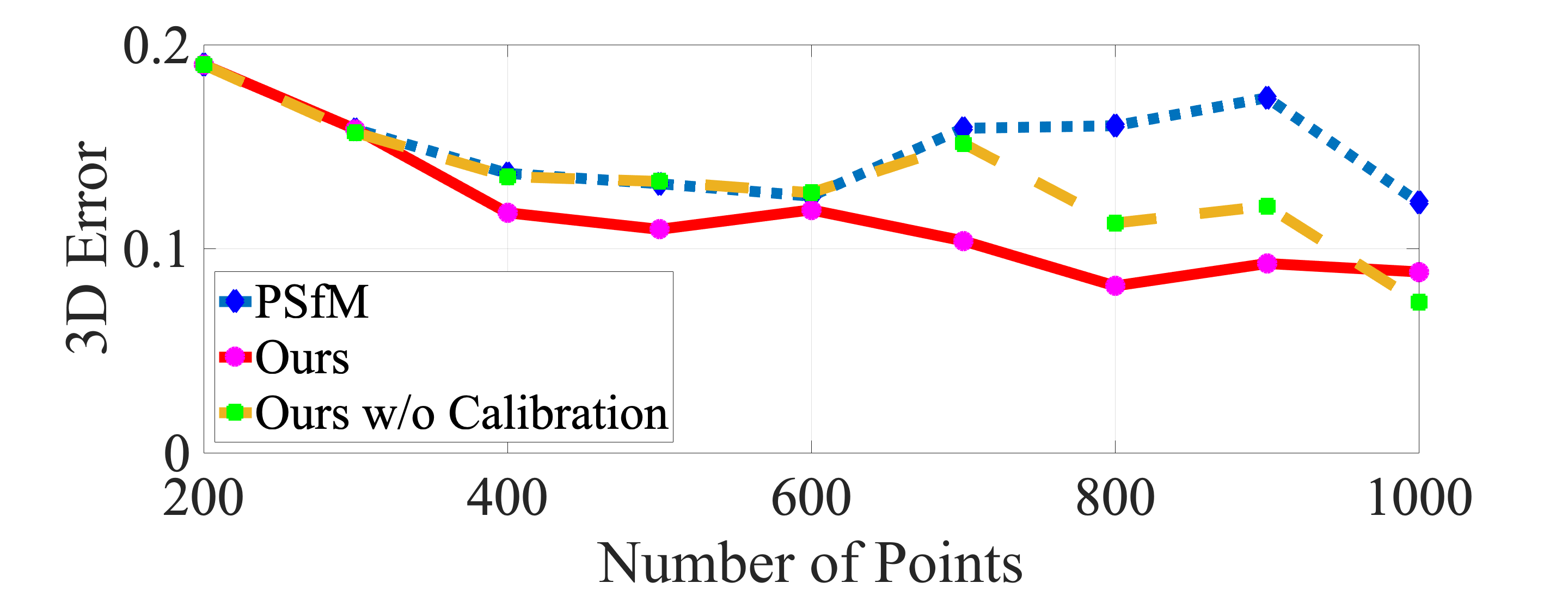}
  \caption{$n=20, \sigma = 0.6\%, \delta = 0.9$, $m$ varies}
  \label{syn_comp_3D:sfig1}
\end{subfigure}
\begin{subfigure}[c]{0.48\textwidth}
  \centering
  \includegraphics[width=\linewidth]{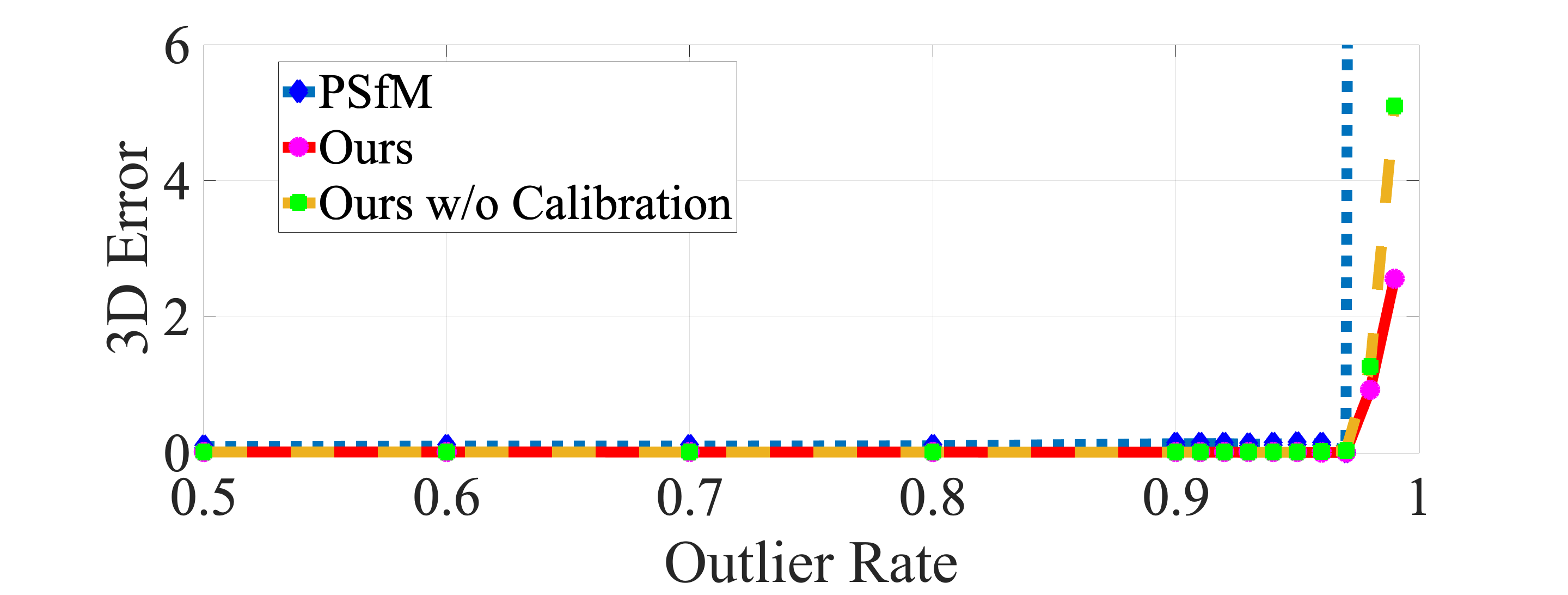}
  \caption{$n=10, m=200, \sigma=0.3\%$, $\delta$ varies}
  \label{syn_comp_3D:sfig2}
\end{subfigure}
\begin{subfigure}[c]{0.48\textwidth}
  \centering
  \includegraphics[width=\linewidth]{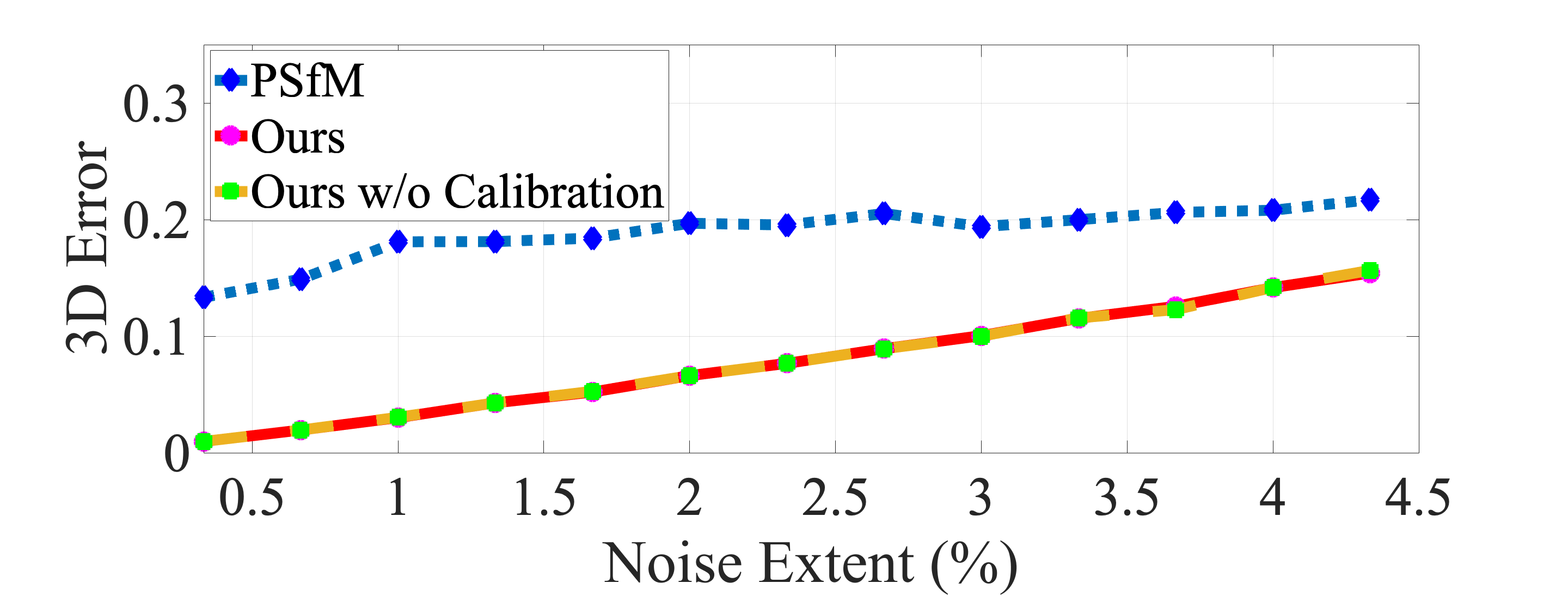}
  \caption{$n=10, m=200, \delta=0.9$, $\sigma$ varies}
  \label{syn_comp_3D:sfig3}
\end{subfigure}
\begin{subfigure}[c]{0.48\textwidth}
  \centering
  \includegraphics[width=\linewidth]{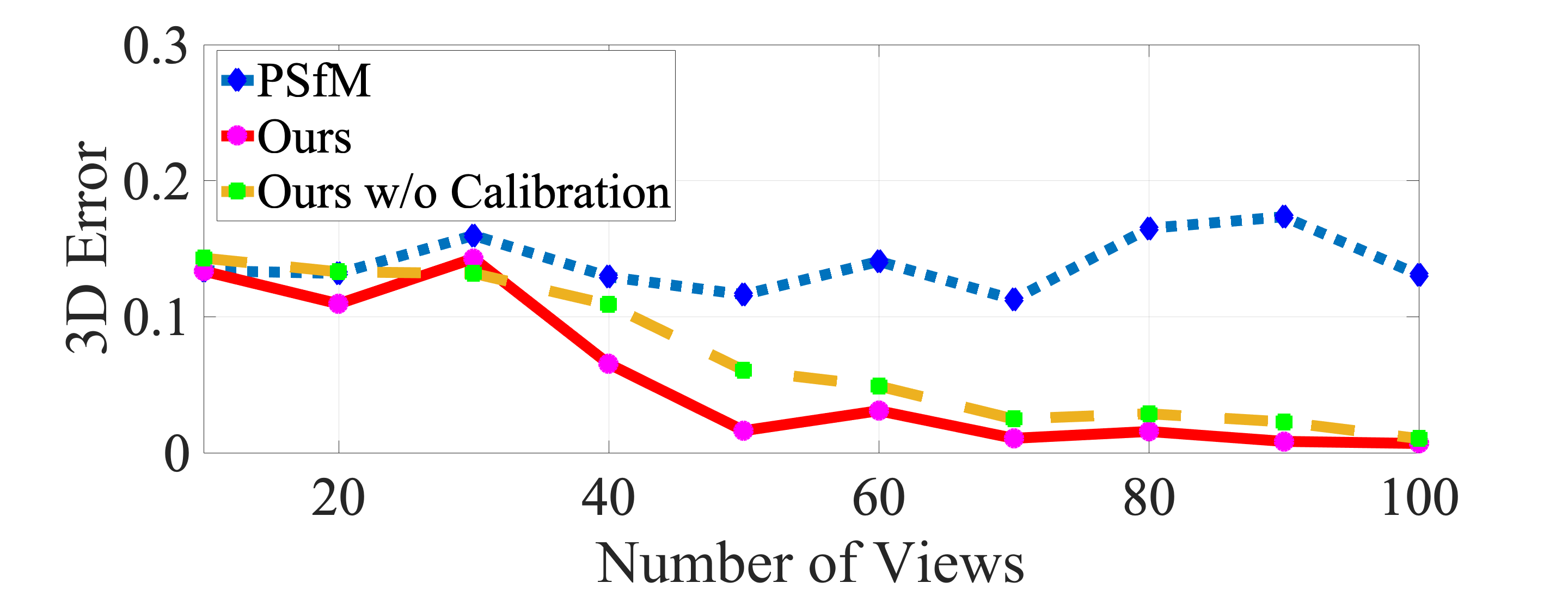}
  \caption{$m=500, \sigma = 0.6\%, \delta = 0.9$, $n$ varies}
  \label{syn_comp_3D:sfig4}
\end{subfigure}
\caption{F1 score, 2D error, and 3D error comparison between SCPSfM model (with and without self-calibration constraints) and the projective structure from motion (PSfM)~\cite{sturm1996factorization}. The reported experiments were conducted with varying: number of points $m$, number of views $n$, outlier rate $\delta$, and noise extent $\delta$.}
\label{fig:syn_comp}
\vspace{-10pt}
\end{figure*}

\section{Experiments}
We evaluated the effectiveness of our model with synthetic and real datasets. On the synthetic data, we measure the effect of different factors: the outlier rate $\delta$, noise extent $\sigma$, number of points $m$ and number of views $n$, in order to show the robustness of our model. We also compare our model with the traditional Projective Structure from Motion (PSfM) method under the same settings. We further perform ablation study of our model with and without the self-calibration support, which is the DAQ loss $\mathcal{L}_{DAQ}$ proposed in Section \ref{sec:SCSFM}. On the real dataset, we combine our model with the state-of-the-art method on projective structure from motion P$^2$SfM \cite{magerand2017practical}. The real experiments also demonstrate that our model quickly rejects most of the outliers. We use our method to reject outliers and obtain the outlier filtered measurement matrix.
The final structure and motion are then recovered providing the outlier filtered measurement matrix into the P$^2$SfM pipeline. This choice is primarily because of the final step of outlier filtering and bundle adjustment offered by the P$^2$SfM. In fact, our experiments show that the proposed method is complimentary to the  P$^2$SfM pipeline. 
\subsection{Synthetic Dataset} \label{sec:syndata}
\subsubsection{Experiments Setup.}
In order to create the synthetic dataset, we randomly generate $m$ number of points and $n$ number of projections with random camera motion. The rotation of the camera is sampled uniformly from the set $[-0.4, 0.4]$ around $xyz$-axis and the translation of the camera is uniformly sampled from $[-1, 1]$. The 3D points are randomly sampled from $[-1, 1]$ along $x$ and $y$ axis and $[2, 4]$ along $z$ axis. In order to guarantee the measurement matrix is meaningful under high outlier rate and high noise extent, the projective depth used in measurement matrix is the ground truth projective depth instead of the estimated projective depth calculated from the fundamental matrix as done in \cite{sturm1996factorization}. The synthetic measurement matrix for exploring the effect of the number of points $m$ and the number of views $n$ has a fixed dimension of $300\times 1000$, but only $3n\times m$ of which consists of valid measurements. The rest of the columns and rows are filled with zeros.
The dimension of the synthetic measurement matrix for exploring the effect of the outlier rate and the noise extent is fixed as $30\times 200$ without zero columns or rows. The outlier correspondences are introduced by exchanging some of the correspondence points and the Gaussian noise is added to all elements in the measurement matrix. We set the hyper parameter $\alpha = 1.0$, $\beta = 1.0$, $t=15.0$ in Eq. (\ref{eq:total_loss}) for all the synthetic experiments. The learning rate is set as 0.001. In order to evaluate the performance of different method, the F1 score, 2D error and 3D error are adopted as the metric for evaluation, following \cite{probst2019unsupervised, magerand2017practical}. 
The F1 score is calculated according to the inlier detection accuracy and the inlier detection recall rate. The 3D error is calculated by $\frac{\norm{X-X_{gt}}}{X_{gt}}$ where $X$ is the reconstructed 3D point and $X_{gt}$ is the ground truth 3D point. The 2D error is the root mean square error in pixel between the reprojected 2D point and the ground truth 2D point and then averaged over all the points. 

\subsubsection{Experimental Results.}
To explore the effect of four factors, number of points $n$, the number of views $m$, the outlier rate $\delta$ and the noise extent $\sigma$ on our model and the traditional projective structure from motion (PSfM) method, we conduct the control variable experiment, whose experimental results are shown in Fig. \ref{fig:syn_comp}.  
From Fig. \ref{syn_comp_F1:sfig1}, Fig. \ref{syn_comp_2D:sfig1} and Fig. \ref{syn_comp_3D:sfig1}, 
it is shown that the performance of all the methods increases with increasing points. 
 It also shows that our model with/without self-calibration part both perform better than the PSfM method.
When the self-calibration constraints are introduces, the performance of our model for fewer points improves further. In Fig. \ref{syn_comp_F1:sfig2}, Fig. \ref{syn_comp_2D:sfig2} and Fig. \ref{syn_comp_3D:sfig2}, PSfM fails in the regime of high outliers, whereas our models with/without calibration constraints provide meaningful results up to $98\% / 97\%$ outlier rate, respectively.
These experiments verify that the use of self-calibration constraints helps to further improve the robustness of our model. In Fig. \ref{syn_comp_F1:sfig3}, Fig. \ref{syn_comp_2D:sfig3} and Fig. \ref{syn_comp_3D:sfig3}, the performance of PSfM drops quickly when noise extent increases, however 
 our model remains very stable in terms of F1 score. It is natural that the 2D and 3D errors of our model 
 increases with increasing noise. Similarly, the performance of all the methods improves with increasing number of views, which can be seen in \ref{syn_comp_F1:sfig4}, Fig. \ref{syn_comp_2D:sfig4} and Fig. \ref{syn_comp_3D:sfig4}. In the same figures, it can also be seen that our model still perform better than PSfM method. As expected, our model with with self calibration constraints is better here again. 
 Overall, our model with/without self-calibration constraints perform better than PSfM with changing number of points/ outlier rate/ noise extent/number of views, when measured in terms of F1 score, 2D and 3D errors. More importantly, our model with self-calibration constraints is consistently better than the one without, ever if it is by a small margin in some cases. For more results and analysis, please refer the supplementary material. 

\subsection{Real Dataset}\label{sec:realdata}
\subsubsection{Experiments Setup.}
\begin{figure*}
     \centering
     \includegraphics[width = \textwidth]{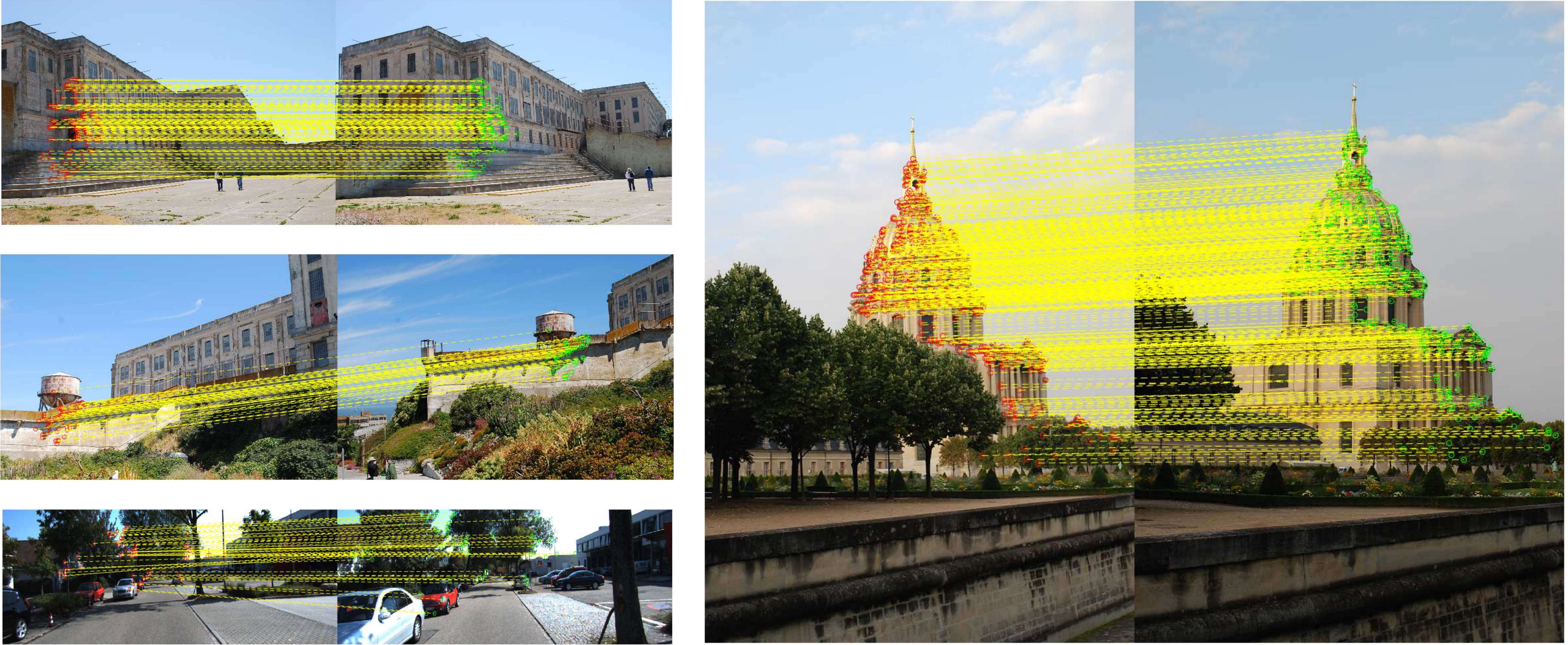}
     \caption{Visualization of the detected correspondence inliers of our model on the Courtyard,  West Side, Dome, and KITTI dataset.}
     \label{fig:visualization}
     \vspace{-10pt}
 \end{figure*}

To verify the effectiveness of our model on the real data, image datasets which cover the multi view images such as Courtyard \cite{olsson2011stable}, West Side \cite{olsson2011stable}, Dome \cite{olsson2011stable} and KITTI \cite{Geiger2013IJRR} are taken. In order to guarantee that there are common correspondences across multiple images, some of the views were rejected. Total number of views, number of correspondence, and the image size are listed in Table \ref{tab:comp_MPSFM_ours}. Due to the limitation of the dataset scale, we report the training results to evaluate our method. Please, note that our method is fully unsupervised. In the whole sequence, except for KITTI, every 10 multi-view images are used to generate one meansurment matrix, i.e., number of views $n = 10$. For KITTI, the number of views $n$ is set as 11. Except for KITTI, the point in 2D images are detected by SIFT \cite{sift:99} and then correspondence is established by the Brute Force Matcher \cite{opencv_library}. For the KITTI dataset, the point and the correspondence matching are taken through the Shi-Tomasi detector \cite{shi1994good} and optical flow \cite{opencv_library}. Due to the unavailability of the projective depth in the real dataset, the projective depth in the measurement matrix is estimated by the fundamental matrix and the epipole following \cite{sturm1996factorization}. In order to evaluate the performance of different methods, the 2D error metric, same as that of synthetic data experiments, is adopted. Besides, the run-time of different methods are also compared. Some qualitative results for matching are shown in Fig.~\ref{fig:visualization}.
\begin{table*}[h]
\setlength{\tabcolsep}{1pt}
\centering
\resizebox{\textwidth}{15mm}
 {
\begin{tabular}{cccc|cc|cc|cc}
  \hline
  \multicolumn{4}{c}{Sequence}&\multicolumn{2}{|c}{Ours+P$^2$SfM}&\multicolumn{2}{|c}{P$^2$SfM\cite{magerand2017practical}}&\multicolumn{2}{|c}{COLMAP\cite{schoenberger2016sfm}}\\  \hline  \hline
  Name&Size&Views&Corrsp.&2D error&Time(s)&2D error&Time(s)&2D error&Time(s)\\
  \hline
   Courtyard \cite{olsson2011stable}&1936$\times$ 1296&21&3000&\textbf{0.2195}&\textbf{16.89}&0.2506&46.74&0.4226&1696\\  \hline  \hline
   West Side \cite{olsson2011stable}&1936$\times$ 1296&97&3000&\textbf{0.2686}&\textbf{28.05}&0.5216&118.93&0.5728&5141\\  \hline  \hline
   Dome \cite{olsson2011stable}&1296$\times$ 1936&81&3000&\textbf{0.1462}&\textbf{21.50}&0.1554&30.49&-&-\\  \hline  \hline
  KITTI \cite{Geiger2013IJRR}&1242$\times$ 375&334&200&\textbf{0.5259}&\textbf{0.08}&\multicolumn{2}{c|}{Not Available}&-&-\\
  \hline   \hline
 \end{tabular}
 }
 \caption{\label{tab:comp_MPSFM_ours} Performance comparison between our method combined with P$^2$SfM, original P$^2$SfM, and COLMAP on the real data. The 2D error and running time are reported for comparisons. The best result is denoted in bold. We run P$^2$SfM and our methods in the very same setup, however experimental setup for COLMAP is different as it also used camera intrinsics and a different pipeline. Therefore, we report the results of COLMAP from \cite{magerand2017practical}. Due to difference in experimental setup, results of COLMAP are not supposed to be compared directly. The latter results are reported here for a general overview.}
\vspace{-10pt}
\end{table*}

\subsubsection{Experimental Results.}
In Table. \ref{tab:comp_MPSFM_ours}, we list the 2D error and the runtime comparisons on the real dataset between our model combined with P$^2$SfM and pure P$^2$SfM. By first taking advantage of our model for rejecting the outliers, the outlier rate of the measurement matrix fed into the P$^2$SfM pipeline becomes much lower. In this way, it becomes easier for the P$^2$SfM pipeline to refine and reconstructed structure and motion.
From Table \ref{tab:comp_MPSFM_ours}, it is proven that the combination of our model with P$^2$SfM outperforms the pure P$^2$SfM according to the 2D error on all the real datasets, 0.2195 v.s. 0.2506, 0.2686 v.s. 0.5216, 0.1462 v.s. 0.1554, respectively. Besides, our model makes the P$^2$SfM much faster for refinement, 16.89s v.s. 46.74s, 28.05s v.s. 118.93s, and 21.50s v.s. 30.49s, respectively. Especially on the West Side dataset, we have 48.5$\%$ improvement on the 2D error and 76.4$\%$ acceleration compared to the pure P$^2$SfM method. Moreover, facing the difficult setting where there are only a few correspondences available on KITTI dataset, the pure P$^2$SfM method does not work and cannot produce the final result. Nevertheless. our model under such setting can work independently to reconstruct with a few correspondences and reach to a meaningful 2D reprojection error of 0.5259. In this setup, our method takes only 0.08s with GPU acceleration.
This further confirms the conclusion we get from synthetic dataset that our model is more stable and robust when fewer number of point correspondences are available, compared to the traditional projective structure from motion method. 
Due to the space limitation, more quantitative and qualitative results on real data are provided in the Supplementary material.

\subsubsection{Camera Intrinsics Prediction.} Since use the DAQ constraints to realize self-calibration in an unsupervised way, we also validate the camera intrinsics prediction by our method. Using the ground truth camera intrinsics $\mathsf{K}$, in the synthetic dataset in Section \ref{sec:syndata}, and the real KITTI dataset, in Section \ref{sec:realdata}, we computed the errors in predicting the focal length. The intrinsics prediction error is calculated through $\frac{\vert f-f_{gt}\vert}{f_{gt}}$, where $f$ is the predicted focal length while the $f_{gt}$ is the ground truth. On an average, our model achieves the accuracy of  $1.67\%$ and $2.32\%$ in predicting the focal length, respectively on synthetic and the real KITTI dataset.

\section{Conclusion}
In this work, we propose the self-calibrating projective structure from motion (SCPSfM) model, which is a unified framework for projective structure from motion and the self-calibration. We have proposed the first unsupervised deep model for solving the projective structure from motion problem, to the best of our knowledge. 
 By exploiting the projective factorization, our model outperforms the traditional projective structure from motion method, both interms of robustness and accuracy. Moreover, when the self-calibration constraints are further exploited, i.e., DAQ constraint, the performance improves further specially in the cases of few views, few points, and high outlier rates. 
  
 The experiments on the synthetic and real datasets verify the effectiveness of our model on recovering  structure and motion together with  self-calibration, while being accurate and extremely robust to outliers. 

{\small
\bibliographystyle{splncs}
\bibliography{egbib}

\begin{thebibliography}{10}

\bibitem{zang}
Zhang, Z.:
\newblock {A flexible new technique for camera calibration}.
\newblock IEEE Transactions on Pattern Analysis and Machine Intelligence
  \textbf{22}(11) (November 2000)  1330--1334

\bibitem{Tsai1987b}
Tsai, R.:
\newblock {A versatile camera calibration technique for high-accuracy 3D
  machine vision metrology using off-the-shelf TV cameras and lenses}.
\newblock IEEE Journal on Robotics and Automation \textbf{3}(4) (August 1987)
  323--344

\bibitem{sift:99}
Lowe, D.G.:
\newblock {Object Recognition from Local Scale-Invariant Features}.
\newblock (1999)  1150--1157

\bibitem{surf:2008}
Bay, T.~Tuytelaars, H., Gool, L.:
\newblock {Surf: Speeded up robust features}.
\newblock \textbf{1} (2006)  404--417

\bibitem{CMS}
Sturm, P.:
\newblock {Critical motion sequences for monocular self-calibration and
  uncalibrated Euclidean reconstruction}.
\newblock Computer Vision and Pattern Recognition, 1997. Proceedings., 1997
  IEEE Computer Society Conference on (1997)  1100--1105

\bibitem{faugeras92}
Faugeras, Q.~Luong, O., Maybank, S.:
\newblock {Camera self-calibration: Theory and experiments}.
\newblock (1992)  321--334

\bibitem{Pollefeys:96}
Pollefeys, L.~Gool, M., Oosterlinck, A.:
\newblock The modulus constraint: a new constraint for self calibration.
\newblock International conference of pattern recognition (1996)  31--42

\bibitem{triggs97}
Triggs, B.:
\newblock {Autocalibration and Absolute Quadric}.
\newblock International Conference on Computer Vision and Pattern Recognition
  (CVPR'97) (1997)  609--614

\bibitem{multiviewGeometry:2003}
Hartley, R., Zisserman, A.:
\newblock Multiple view geometry.
\newblock Cambridge University Press (2003)

\bibitem{KoenderinkandvanDoorn}
Koenderink, J., van Doorn, A.:
\newblock {Affine structure from motion.}
\newblock Journal of the Optical Society of America. A, Optics and image
  science \textbf{8}(2) (February 1991)  377--385

\bibitem{faugeras95}
Faugeras, O.:
\newblock {Stratification of three-dimensional vision: projective, affine, and
  metric representations: errata}.
\newblock J. Opt. Soc. Am. A \textbf{12}(7) (July 1995)  1606+

\bibitem{loung96}
Luong, Vieville, T.:
\newblock {Canonical Representations for the Geometries of Multiple Projective
  Views}.
\newblock Computer Vision and Image Understanding \textbf{64}(2) (September
  1996)  193--229

\bibitem{adlakha2019quarch}
Adlakha, D., Habed, A., Morbidi, F., Demonceaux, C., Mathelin, M.d.:
\newblock Quarch: A new quasi-affine reconstruction stratum from vague relative
  camera orientation knowledge.
\newblock In: Proceedings of the IEEE International Conference on Computer
  Vision. (2019)  1082--1090

\bibitem{Liebowitz}
Liebowitz, D., Zisserman, A.:
\newblock {Combining scene and auto-calibration constraints}.
\newblock Computer Vision, 1999. The Proceedings of the Seventh IEEE
  International Conference on \textbf{1} (1999)  293--300 vol.1

\bibitem{pStrum99}
Sturm, P., Maybank, S.:
\newblock {On Plane-Based Camera Calibration: A General Algorithm,
  Singularities, Applications} (1999)

\bibitem{Faugeras:95}
Faugeras, G.~Laveau, S.R.L.C.O., Zeller, C.:
\newblock 3d reconstruction of urban scene from sequence of images.
\newblock Technical report, INRIA (1995)

\bibitem{habed2014efficient}
Habed, A., Pani~Paudel, D., Demonceaux, C., Fofi, D.:
\newblock Efficient pruning lmi conditions for branch-and-prune rank and
  chirality-constrained estimation of the dual absolute quadric.
\newblock In: Proceedings of the IEEE Conference on Computer Vision and Pattern
  Recognition. (2014)  493--500

\bibitem{rank3:2007}
Chandraker, M.~Agarwal, S.K.F.N.D., Kriegman, D.:
\newblock Practical autocalibration.
\newblock Computer Vision and Pattern Recognition (2007)

\bibitem{Pollefeys-98}
Pollefeys, L.~Gool, M., Koch, M.:
\newblock {Self-Calibration and Metric Reconstruction in Spite of Varying and
  Unknown Internal Camera Parameters}.
\newblock (1998)  90--95

\bibitem{Nister:04}
Nister, D.:
\newblock Untwisting a projective reconstruction.
\newblock International Journal of Computer Vision (November,2004)  165--183

\bibitem{Gherardi:2010}
Gherardi, R., Fusiello, A.:
\newblock Practical autocalibration.
\newblock European Conference on Computer Vision (2010)

\bibitem{pStrumFactorization:1996}
Strum, P., Triggs, B.:
\newblock A factorization based algorithm for multi-image projective structure
  and motion.
\newblock European Conference on Computer Vision, Cambridge, England (April,
  1996)  709--720

\bibitem{gurdjos2009dual}
Gurdjos, P., Bartoli, A., Sturm, P.:
\newblock Is dual linear self-calibration artificially ambiguous?
\newblock In: 2009 IEEE 12th International Conference on Computer Vision, IEEE
  (2009)  88--95

\bibitem{Schaffalitzky2002}
Schaffalitzky, F., Zisserman, A.:
\newblock {Multi-view Matching for Unordered Image Sets, or "How Do I Organize
  My Holiday Snaps?"}.
\newblock (2002)  414--431

\bibitem{montserrat2019multi}
Montserrat, D.M., Chen, J., Lin, Q., Allebach, J.P., Delp, E.J.:
\newblock Multi-view matching network for 6d pose estimation.
\newblock arXiv preprint arXiv:1911.12330 (2019)

\bibitem{schonberger2016pixelwise}
Sch{\"o}nberger, J.L., Zheng, E., Frahm, J.M., Pollefeys, M.:
\newblock Pixelwise view selection for unstructured multi-view stereo.
\newblock In: European Conference on Computer Vision, Springer (2016)  501--518

\bibitem{serlin2019distributed}
Serlin, Z., Yang, G., Sookraj, B., Belta, C., Tron, R.:
\newblock Distributed and consistent multi-image feature matching via
  quickmatch.
\newblock arXiv preprint arXiv:1910.13317 (2019)

\bibitem{mahamud2001provably}
Mahamud, S., Hebert, M., Omori, Y., Ponce, J.:
\newblock Provably-convergent iterative methods for projective structure from
  motion.
\newblock In: Proceedings of the 2001 IEEE Computer Society Conference on
  Computer Vision and Pattern Recognition. CVPR 2001. Volume~1., IEEE (2001)
  I--I

\bibitem{hartley2003powerfactorization}
Hartley, R., Schaffalitzky, F.:
\newblock Powerfactorization: 3d reconstruction with missing or uncertain data.
\newblock In: Australia-Japan advanced workshop on computer vision. Volume~74.
  (2003)  76--85

\bibitem{oliensis2007iterative}
Oliensis, J., Hartley, R.:
\newblock Iterative extensions of the sturm/triggs algorithm: Convergence and
  nonconvergence.
\newblock IEEE Transactions on Pattern Analysis and Machine Intelligence
  \textbf{29}(12) (2007)  2217--2233

\bibitem{dai2010element}
Dai, Y., Li, H., He, M.:
\newblock Element-wise factorization for n-view projective reconstruction.
\newblock In: European Conference on Computer Vision, Springer (2010)  396--409

\bibitem{magerand2017practical}
Magerand, L., Del~Bue, A.:
\newblock Practical projective structure from motion (p2sfm).
\newblock In: ICCV. (2017)

\bibitem{zhou2017unsupervised}
Zhou, T., Brown, M., Snavely, N., Lowe, D.G.:
\newblock Unsupervised learning of depth and ego-motion from video.
\newblock In: Proceedings of the IEEE Conference on Computer Vision and Pattern
  Recognition. (2017)  1851--1858

\bibitem{godard2019digging}
Godard, C., Mac~Aodha, O., Firman, M., Brostow, G.J.:
\newblock Digging into self-supervised monocular depth estimation.
\newblock In: Proceedings of the IEEE International Conference on Computer
  Vision. (2019)  3828--3838

\bibitem{chen2019self}
Chen, Y., Schmid, C., Sminchisescu, C.:
\newblock Self-supervised learning with geometric constraints in monocular
  video: Connecting flow, depth, and camera.
\newblock In: Proceedings of the IEEE International Conference on Computer
  Vision. (2019)  7063--7072

\bibitem{pedra2013camera}
Pedra, A.V.B.M., Mendon{\c{c}}a, M., Finocchio, M.A.F., de~Arruda, L.V.R.,
  Castanho, J.E.C.:
\newblock Camera calibration using detection and neural networks.
\newblock IFAC Proceedings Volumes \textbf{46}(7) (2013)  245--250

\bibitem{bogdan2018deepcalib}
Bogdan, O., Eckstein, V., Rameau, F., Bazin, J.C.:
\newblock Deepcalib: a deep learning approach for automatic intrinsic
  calibration of wide field-of-view cameras.
\newblock In: Proceedings of the 15th ACM SIGGRAPH European Conference on
  Visual Media Production. (2018)  1--10

\bibitem{hold2018perceptual}
Hold-Geoffroy, Y., Sunkavalli, K., Eisenmann, J., Fisher, M., Gambaretto, E.,
  Hadap, S., Lalonde, J.F.:
\newblock A perceptual measure for deep single image camera calibration.
\newblock In: Proceedings of the IEEE Conference on Computer Vision and Pattern
  Recognition. (2018)  2354--2363

\bibitem{gordon2019depth}
Gordon, A., Li, H., Jonschkowski, R., Angelova, A.:
\newblock Depth from videos in the wild: Unsupervised monocular depth learning
  from unknown cameras.
\newblock In: Proceedings of the IEEE International Conference on Computer
  Vision. (2019)  8977--8986

\bibitem{zhuang2019degeneracy}
Zhuang, B., Tran, Q.H., Ji, P., Lee, G.H., Cheong, L.F., Chandraker, M.:
\newblock Degeneracy in self-calibration revisited and a deep learning solution
  for uncalibrated slam.
\newblock arXiv preprint arXiv:1907.13185 (2019)

\bibitem{ranftl2018deep}
Ranftl, R., Koltun, V.:
\newblock Deep fundamental matrix estimation.
\newblock In: Proceedings of the European Conference on Computer Vision (ECCV).
  (2018)  284--299

\bibitem{probst2019unsupervised}
Probst, T., Paudel, D.P., Chhatkuli, A., Gool, L.V.:
\newblock Unsupervised learning of consensus maximization for 3d vision
  problems.
\newblock In: Proceedings of the IEEE Conference on Computer Vision and Pattern
  Recognition. (2019)  929--938

\bibitem{brachmann2019neural}
Brachmann, E., Rother, C.:
\newblock Neural-guided ransac: Learning where to sample model hypotheses.
\newblock In: Proceedings of the IEEE International Conference on Computer
  Vision. (2019)  4322--4331

\bibitem{brachmann2017dsac}
Brachmann, E., Krull, A., Nowozin, S., Shotton, J., Michel, F., Gumhold, S.,
  Rother, C.:
\newblock Dsac-differentiable ransac for camera localization.
\newblock In: Proceedings of the IEEE Conference on Computer Vision and Pattern
  Recognition. (2017)  6684--6692

\bibitem{sturm1996factorization}
Sturm, P., Triggs, B.:
\newblock A factorization based algorithm for multi-image projective structure
  and motion.
\newblock In: European conference on computer vision (ECCV). (1996)

\bibitem{qi2017pointnet}
Qi, C.R., Su, H., Mo, K., Guibas, L.J.:
\newblock Pointnet: Deep learning on point sets for 3d classification and
  segmentation.
\newblock In: CVPR. (2017)

\bibitem{NEURIPS2019_9015}
Paszke, A., Gross, S., Massa, F., Lerer, A., Bradbury, J., Chanan, G., Killeen,
  T., Lin, Z., Gimelshein, N., Antiga, L., Desmaison, A., Kopf, A., Yang, E.,
  DeVito, Z., Raison, M., Tejani, A., Chilamkurthy, S., Steiner, B., Fang, L.,
  Bai, J., Chintala, S.:
\newblock Pytorch: An imperative style, high-performance deep learning library.
\newblock In: NIPS.
\newblock (2019)

\bibitem{kingma2014adam}
Kingma, D.P., Ba, J.:
\newblock Adam: A method for stochastic optimization.
\newblock ICLR (2014)

\bibitem{olsson2011stable}
Olsson, C., Enqvist, O.:
\newblock Stable structure from motion for unordered image collections.
\newblock In: Scandinavian Conference on Image Analysis. (2011)

\bibitem{Geiger2013IJRR}
Geiger, A., Lenz, P., Stiller, C., Urtasun, R.:
\newblock Vision meets robotics: The kitti dataset.
\newblock International Journal of Robotics Research (IJRR) (2013)

\bibitem{opencv_library}
Bradski, G.:
\newblock {The OpenCV Library}.
\newblock Dr. Dobb's Journal of Software Tools (2000)

\bibitem{shi1994good}
Shi, J.,  et~al.:
\newblock Good features to track.
\newblock In: CVPR. (1994)

\bibitem{schoenberger2016sfm}
Sch\"{o}nberger, J.L., Frahm, J.M.:
\newblock Structure-from-motion revisited.
\newblock In: Conference on Computer Vision and Pattern Recognition (CVPR).
  (2016)

\end{thebibliography}
}
\pagestyle{headings}
\title{Supplementary Material for \\ Self-Calibration Supported Robust Projective Structure-from-Motion}

\titlerunning{Self-Calibration Supported Robust Projective SfM}
\authorrunning{Rui Gong, Danda Pani Paudel, Ajad Chhatkuli, and Luc Van Gool}
\author{Rui Gong\Mark{1}, Danda Pani Paudel\Mark{1}, Ajad Chhatkuli\Mark{1}, and Luc Van Gool\Mark{1,2}}
\institute{	\Mark{1}\,Computer Vision Lab, ETH Z\"urich, Switzerland \\
		\Mark{2}\,VISICS, ESAT/PSI, KU Leuven, Belgium \\
	\email{\{gongr,paudel,ajad.chhatkuli,vangool\}@vision.ee.ethz.ch}}
\maketitle

In this Supplementary document, we provide additional information with:
\begin{itemize}
    \item More quantitative results and analysis on the the synthetic dataset
    \item More qualitative and quantitative results and analysis on the real dataset 
\end{itemize}

\section{Additional Results on Synthetic Dataset}
In Section 5.1 of the main paper, the quantitative comparison between our self-calibration supported robust projective structure-from-motion model (SCPSfM) and the traditional projective structure-from-motion (PSfM) method is conducted under the different settings of: number of points $n$, number of views $m$, outlier rate $\delta$ and noise extent $\sigma$. In Fig. 2 of the main paper, the comparison is shown to prove the advantage of our model with calibration constraint compared with our model without calibration and PSfM method. In Fig. 2, it is shown that our model with/without self-calibration constraint both outperforms the traditional PSfM method under all the settings. Moreover, our model with self-calibration constraint consistently performs better than our model without self-calibration constraint, which can be seen from the obvious margin between the curves of with and without calibration constraint in Fig. 2 of the main paper. The margin can be observed in all the cases when varying the number of points $n$, the number of views $m$, the outlier rate $\delta$ in Fig. 2 of the main paper. But due to low range used to explore the effect of noise extent $\sigma$, the curve of our model with calibration constraint only shows small improvement compared to the curve without calibration constraint when varying noise extent (ref. Fig. 2(c)(g)(k) in the main paper). In order to show the robustness and benefit of our model from the self-calibration constraint when facing different extent of noise, we provide more experimental results on the synthetic dataset here.
We further increase the noise extent $\sigma$ to higher noise extent compared with the experiment in the main paper. The results of the experiments are plotted in Fig. \ref{fig:syn_comp}, which shows that our model with the self-calibration constraint is more robust and performs much better especially when facing high noise condition. It is notable that our model with self-calibration constraint can stand the $11\%$ noise while the PSfM method and our model without calibration constraint does not work at all under such high noise. It further verifies the robustness and advantage of our SCPSfM model profiting from the self-calibration constraint. 
\section{Additional Results on Real Dataset}
In Section 5.2 of the main paper, we provide the quantitative performance comparison between our model combined with P$^{2}$SfM and pure P$^{2}$SfM on the real dataset. The Table 1 of the main paper shows the advantage of our model for accelerating and reducing the error of the P$^{2}$SfM. In order to further verify the conclusion that we draw, we here provide more comparison results on additional real datasets, which are listed in Table \ref{tab:comp_MPSFM_ours}. The experiment setup is exactly the same as done in Section 5.2 of the main paper. From Table \ref{tab:comp_MPSFM_ours}, it is shown that the combination of our model with P$^{2}$SfM method outperforms the pure P$^2$SfM method according to 2D error, 0.2387 v.s. 0.3187, 0.1576 v.s. 0.1665, 0.2106 v.s. 0.4261 and 0.1596 v.s. 0.1912. Moreover, the speed of the P$^2$SfM is also highly improved profiting from our model, 23.43s v.s. 45.41s, 24.05s v.s. 35.99s, 18.61s v.s. 72.52s and 28.75s v.s. 44.78s. It further proves the benefit of our model on the accuracy and speed of the projective structure-from-motion. Besides the quantitative results, Fig. \ref{supp:visualization} provides the qualitative results of detected correspondence inliers of our method combined with P$^2$SfM on the additional real datasets.

\begin{table*}[h]
\setlength{\tabcolsep}{3pt}
\centering
\resizebox{\textwidth}{15mm}
 {
\begin{tabular}{cccc|cc|cc}
  \hline
  \multicolumn{4}{c}{Sequence}&\multicolumn{2}{|c}{Ours+P$^2$SfM}&\multicolumn{2}{|c}{P$^2$SfM\cite{magerand2017practical}}\\  \hline  \hline
  Name&Size&Views&Corrsp.&2D error&Time(s)&2D error&Time(s)\\
  \hline
   De Guerre \cite{olsson2011stable}&1296$\times$ 1936&20&2000&\textbf{0.2387}&\textbf{23.43}&0.3187&45.41\\  \hline  \hline
   Lund Cathedral \cite{olsson2011stable}&1296$\times$ 1936&50&3000&\textbf{0.1576}&\textbf{24.05}&0.1665&35.99\\  \hline  \hline
   UWO \cite{olsson2011stable}&1296$\times$ 1936&20&3000&\textbf{0.2106}&\textbf{18.61}&0.4261&72.52\\  \hline  \hline
  Water Tower \cite{olsson2011stable}&1296$\times$ 1936&170&3000&\textbf{0.1596}&\textbf{28.75}&0.1912&44.78\\
  \hline   \hline
 \end{tabular}
 }
 \caption{\label{tab:comp_MPSFM_ours} Performance comparison between our method combined with P$^2$SfM and the original P$^2$SfM on the real data. The 2D error and running time are reported for comparisons. The best values for the 2D error and time are in bold. We run P$^2$SfM and our methods in the very same setup.}
\end{table*}

\begin{figure*}
\centering
\begin{subfigure}[c]{0.8\textwidth}
  \centering
  \includegraphics[width=\linewidth]{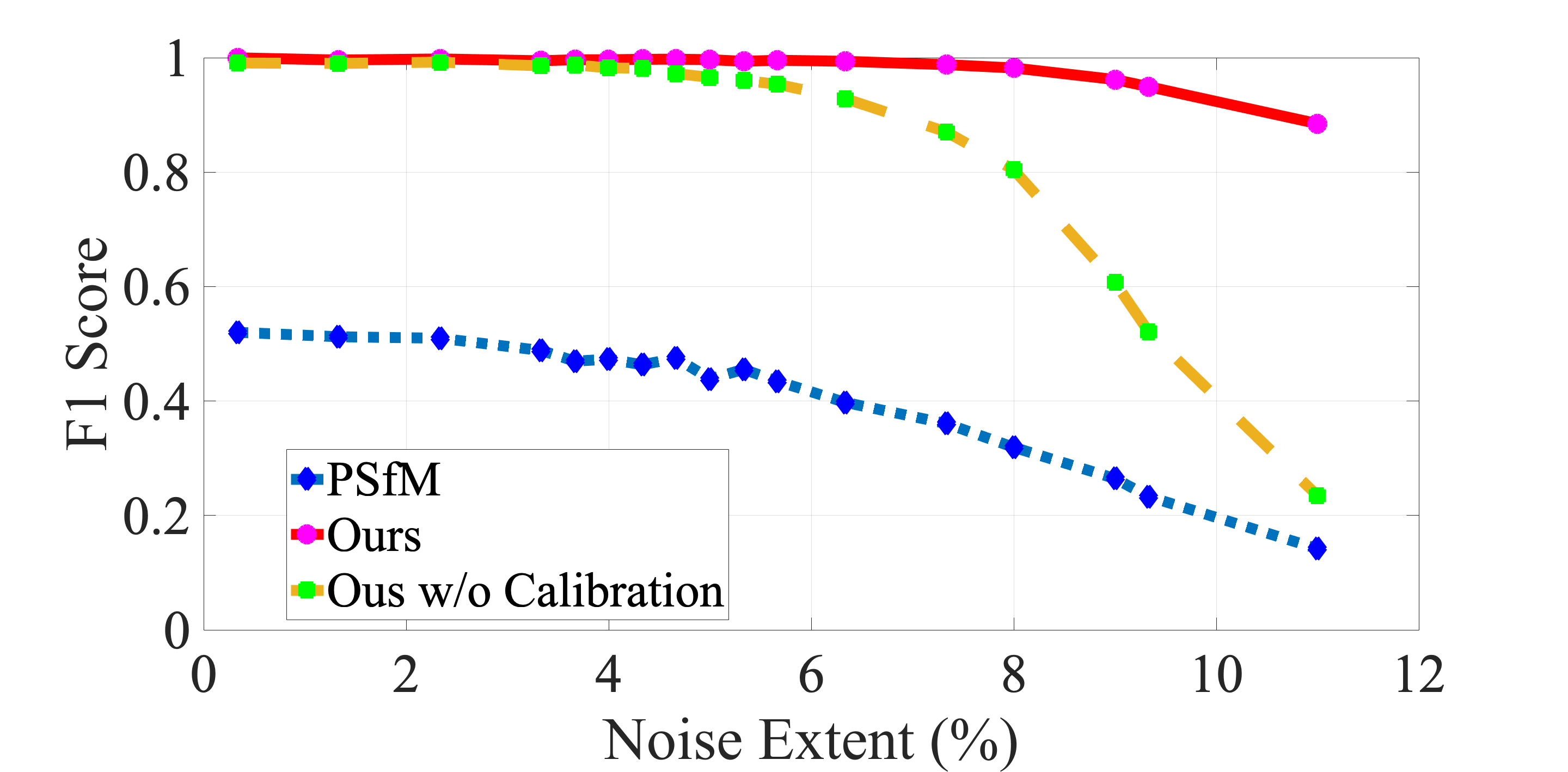}
  \caption{$n=10, m=200, \delta=0.96$, $\sigma$ varies}
  \label{syn_comp_F1score}
\end{subfigure}
\begin{subfigure}[c]{0.8\textwidth}
  \centering
  \includegraphics[width=\linewidth]{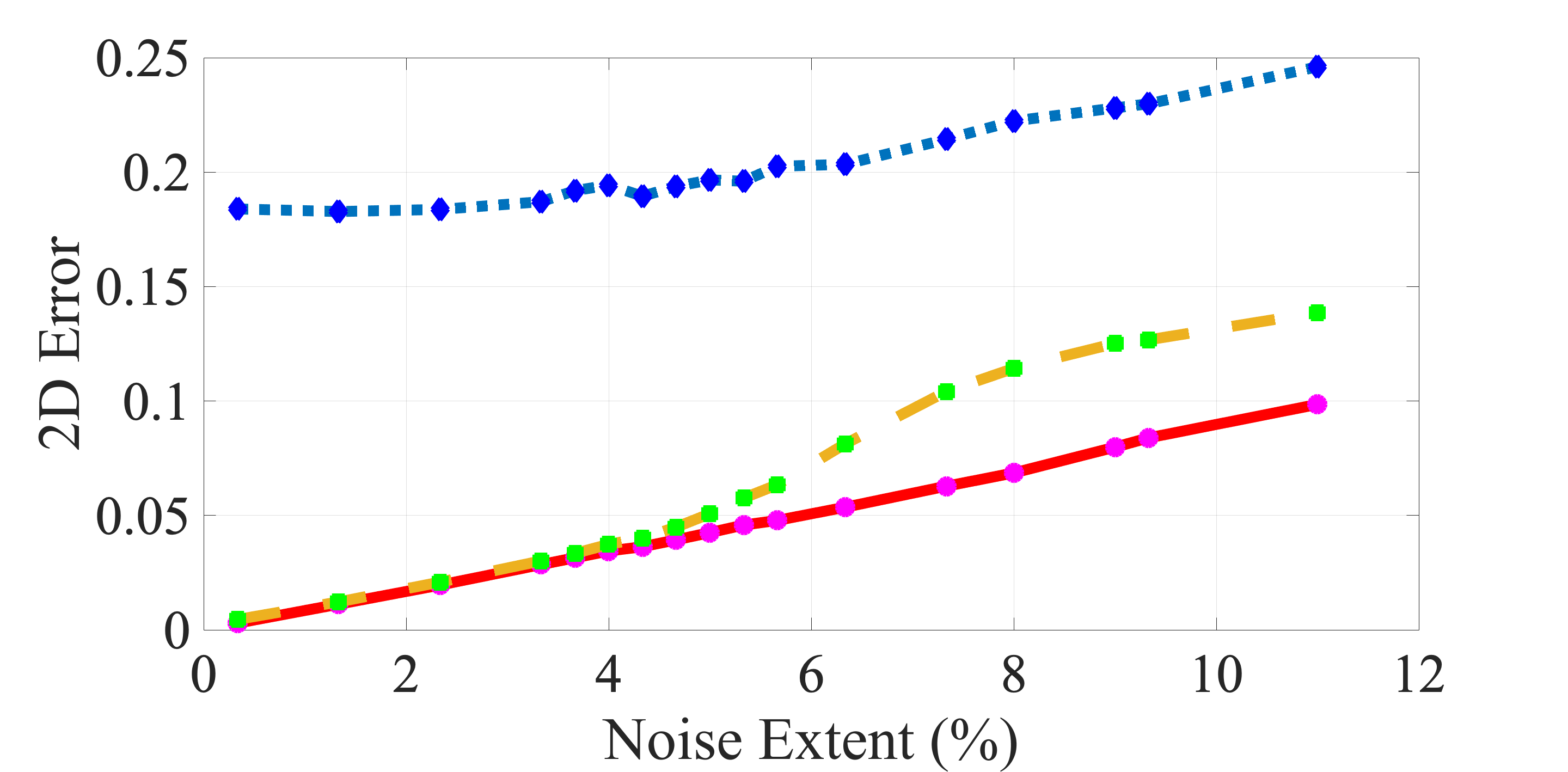}
  \caption{$n=10, m=200, \delta=0.96$, $\sigma$ varies}
  \label{syn_comp_2DError}
\end{subfigure}
\begin{subfigure}[c]{0.8\textwidth}
  \centering
  \includegraphics[width=\linewidth]{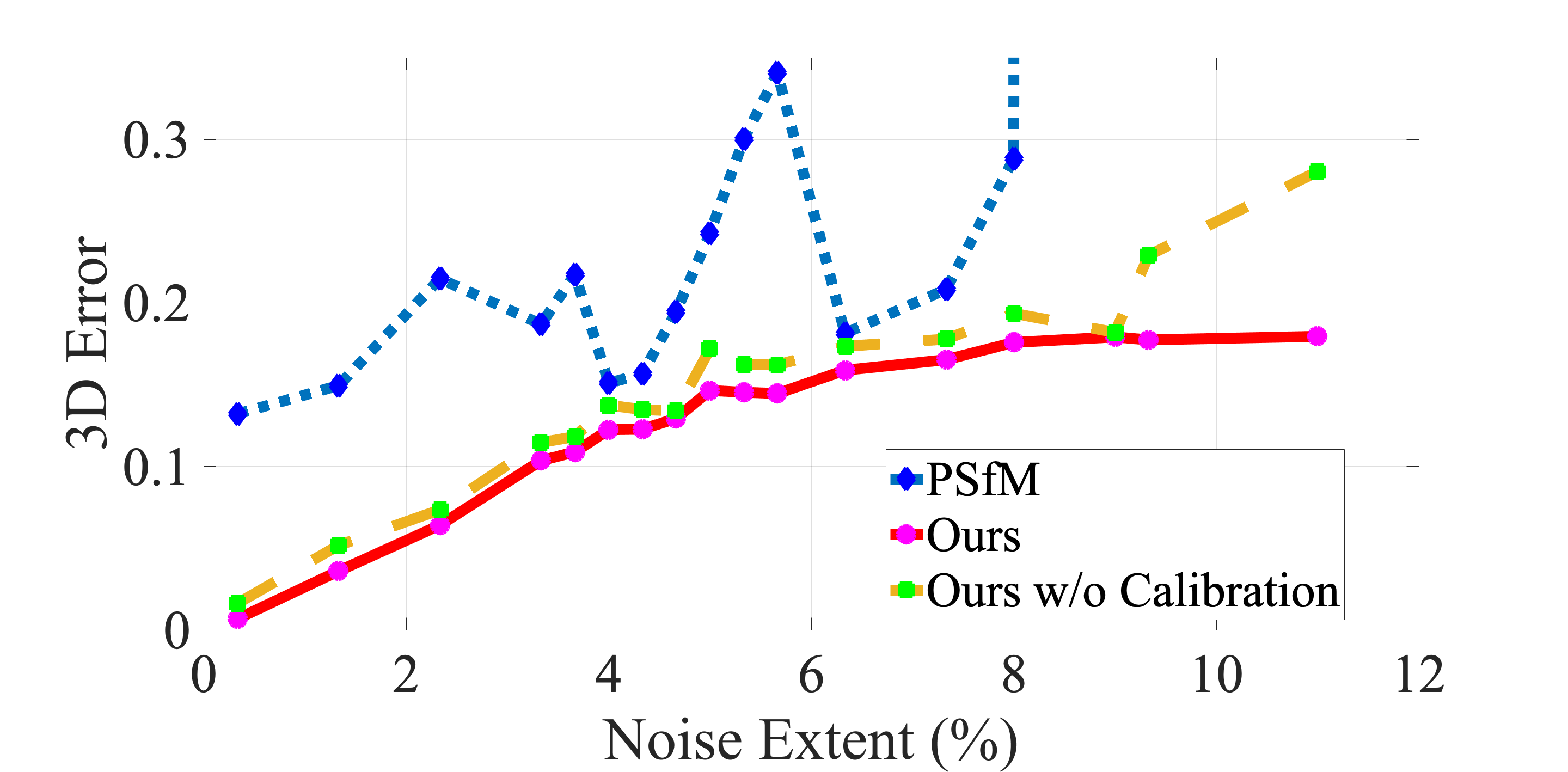}
  \caption{$n=10, m=200, \delta=0.96$, $\sigma$ varies}
  \label{syn_comp_3DError}
\end{subfigure}
\caption{F1 score, 2D error, and 3D error comparison between SCPSfM model (with and without self-calibration constraints) and the projective structure-from-motion (PSfM)~\cite{sturm1996factorization}. The reported experiments were conducted with varying noise extent $\sigma$ and fixing number of points $m$, number of views $n$, outlier rate $\delta$.}
\label{fig:syn_comp}
\end{figure*}

\begin{figure*}
     \centering
     \includegraphics[width = \textwidth]{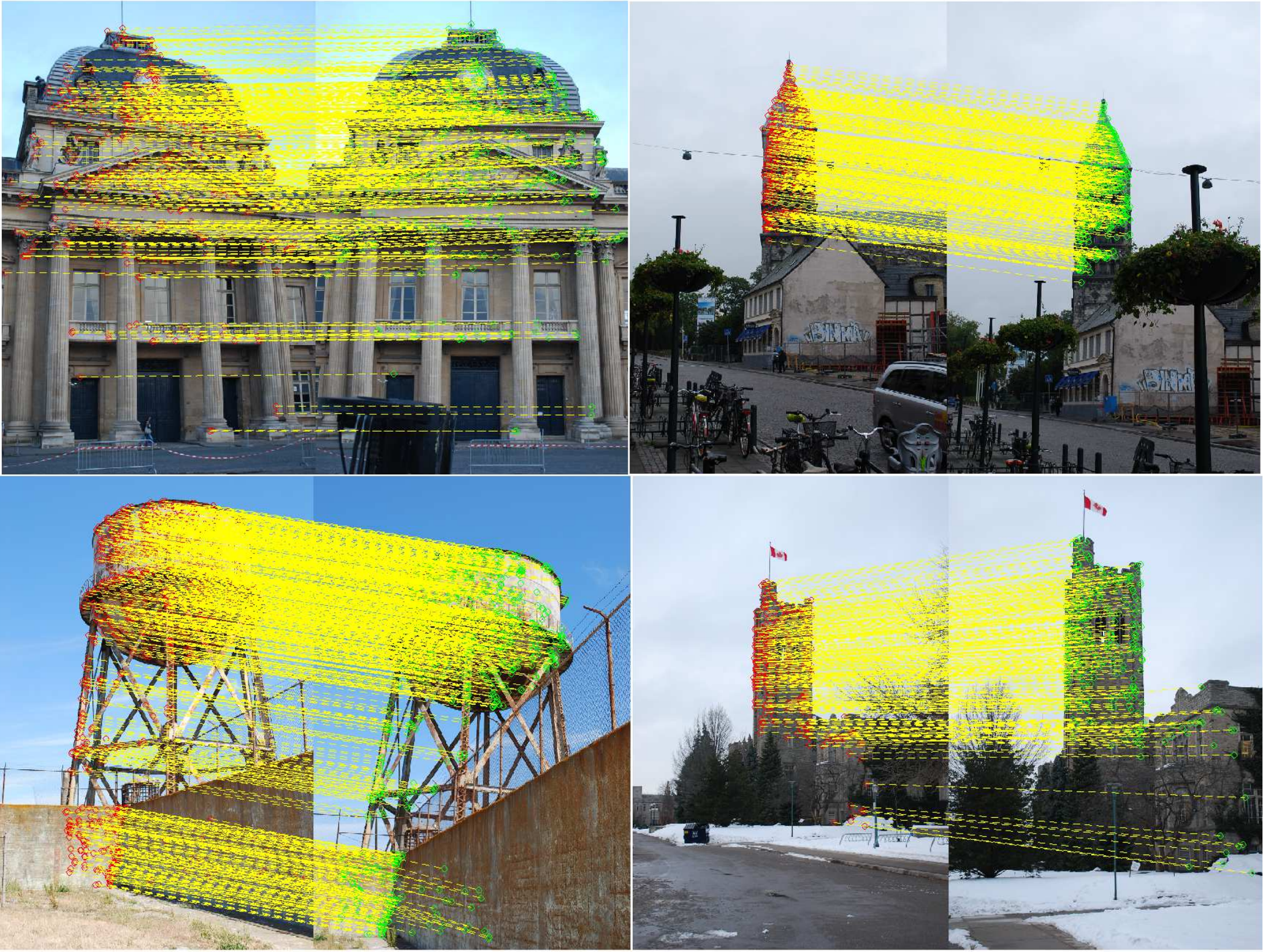}
     \caption{Visualization of the detected correspondence inliers of our model combined with P$^{2}$SfM on the De Guerre, Lund Cathedral, UWO, and Water Tower dataset.}
     \label{supp:visualization}
 \end{figure*}

\end{document}